\documentclass{article}

    \PassOptionsToPackage{numbers, compress}{natbib}

\usepackage[preprint]{neurips_2023}

\usepackage[utf8]{inputenc} 
\usepackage[T1]{fontenc}    
\usepackage{hyperref}       
\usepackage{url}            
\usepackage{booktabs}       
\usepackage{amsfonts}       
\usepackage{nicefrac}       
\usepackage{microtype}      
\usepackage{xcolor}         
\usepackage{graphicx}
\usepackage{layouts}
\usepackage{subfig}
\usepackage{wrapfig}
\usepackage{enumitem}
\usepackage{amsmath}
\usepackage{amssymb}
\usepackage{mathtools}
\usepackage{amsthm}
\usepackage{titlesec}
\usepackage{placeins}
\usepackage{array}
\titlespacing{\paragraph}{0pt}{0pt}{0.5em}
\titlespacing{\section}{0pt}{4pt}{0pt}
\titlespacing{\subsection}{0pt}{0pt}{0pt}
\title{Probing clustering in neural network representations}

\author{%
  Thao Nguyen\thanks{Work done while at Google}\\
  University of Washington\\
  \And
  Simon Kornblith\footnotemark[1]\\
  Anthropic\\
}

\begin{document}

\maketitle
\begin{abstract}
Neural network representations contain structure beyond what was present in the training labels. For instance, representations of images that are visually or semantically similar tend to lie closer to each other than to dissimilar images, regardless of their labels. 
Clustering these representations can thus provide insights into dataset properties as well as the network internals. 
In this work, we study how the many design choices involved in neural network training affect the clusters formed in the hidden representations. To do so, we establish an evaluation setup based on the BREEDS hierarchy \cite{santurkar2020breeds}, for the task of subclass clustering after training models with only superclass information.
We isolate the training dataset and architecture as important factors affecting clusterability. Datasets with labeled classes consisting of unrelated subclasses yield much better clusterability than those following a natural hierarchy. When using pretrained models to cluster representations on downstream datasets, models pretrained on subclass labels provide better clusterability than models pretrained on superclass labels, but only when there is a high degree of domain overlap between the pretraining and downstream data. Architecturally, we find that normalization strategies affect which layers yield the best clustering performance, and, surprisingly, Vision Transformers attain lower subclass clusterability than ResNets. 
\end{abstract}
\section{Introduction}
Today, it costs about 3000x less to train a computer vision model to AlexNet-level accuracy than it did a decade ago. To make use of the power afforded by new microprocessor technologies and more efficient architectures and training techniques, researchers have constructed ever-larger image datasets. These datasets are so large that it is impossible to understand what they contain without the assistance of machine learning models. If one spent a second looking at each image, it would take more than a lifetime to look at the entire JFT-4B dataset, on which today’s top-performing ImageNet models are pre-trained.

An easy way to distill a large dataset to a representative set of images is by clustering representations.
By embedding images in the representation space of a pre-trained model and then examining only a few images from each cluster, one can obtain an understanding of what the dataset contains without laboriously inspecting individual images.  For example, \citet{oakden2020hidden} show that performing k-means clustering in the embedding space of a network trained on the CXR-14 chest X-ray dataset reveals a large cluster of cases with chest drains in the pneuomothorax class. Since a chest drain is inserted to treat pneuomothorax, a model that simply detects the presence or absence of a chest drain to identify pneuomothorax has no diagnostic value. Given this observation, the authors highlight the importance of detecting hidden stratification: when the labels used to train and evaluate machine learning models do not sufficiently describe meaningful variations within the population.

Despite the utility of clustering for understanding dataset composition, the study of factors that affect clustering of representations in pre-trained models has received relatively little attention in the deep learning literature. Instead, significant research effort has been devoted to the design of novel deep algorithms to cluster images in an unsupervised fashion, where networks are trained solely for the purpose of clustering. By contrast, we analyze how various aspects of the standard training procedure can affect the cluster quality in representation spaces of the resulting model. We believe that our setup closely resembles how clustering methods are likely to be applied to large datasets in practice, since training new models on these datasets requires significant computational resources.

Beyond offering insights on the data, clustering can also provide a novel lens through which to view neural network representations. Previous work has proposed to understand properties of hidden representations by measuring their similarity~\cite{raghu2017svcca,kornblith2019similarity,nguyen2021do,williams2021generalized,ding2021grounding}, compatibility~\cite{lenc2015understanding,bansal2021revisiting,csiszarik2021similarity}, or through the performance of simple classifiers (``linear probes'') trained on them~\cite{alain2017understanding}. We demonstrate that clustering is a useful addition to this toolbox. Compared to the aforementioned approaches, clustering is less sensitive to distortions of global representational structure, and because clusters can be visualized directly, differences in representational properties revealed through clustering are potentially more interpretable than a scalar measure of similarity or accuracy.

In our experiments, we leverage the BREEDS hierarchy~\citep{santurkar2020breeds} to create a realistic setup where models are trained on superclass labels, but each superclass contains subclasses for which labels are not observed at training time. We cluster representations within each superclass and measure the accuracy with which we can recover the original subclasses. We also measure clusterability of representations of downstream data that differs in distribution from the initial training set. Using this setup, we uncover several previously unrecognized phenomena. Our findings are as follows:
\begin{itemize}[leftmargin=6pt,topsep=0pt,itemsep=0pt,parsep=4pt]
\item When dataset classes are generated by merging unrelated subclasses, representations of images belonging to these subclasses tend to form separate clusters. When subclasses follow a natural hierarchy, separating them in the representation space can be significantly harder.
\item When using a pre-trained model to cluster image embeddings from downstream datasets that have little overlap with the training set, pre-training on subclass labels yields only marginally better clusterability of downstream classes than pre-training on superclass labels.
\item Clusterability depends on architecture and normalization strategy. Representations of Vision Transformers (ViTs) are less clusterable than representations of similarly peforming ResNets.
\item Objectives and layers that offer the highest linear probe accuracy and transferability to downstream tasks do not necessarily offer the best clusterability.
\item Given the same training set and loss function, different training runs yield similar clustering performance when compared to ground-truth subclasses. However, the actual cluster assignments are highly inconsistent across runs.
\end{itemize}

All in all, these findings help us put the approach of clustering neural network representations, especially in the presence of hidden stratification, on a more solid empirical footing.
\section{Related work}
Clustering of embeddings has been commonly used in various machine learning methods and applications. 
\citet{oakden2020hidden,sohoni2020no} investigate the use of clustering to identify ``hidden stratification,'' i.e., subclasses on which models achieve different performance; \citet{sohoni2020no} further show that retraining on the clusters found using distributionally robust optimization can improve worst-case generalization error.  \citet{birodkar2019semantic,sorscher2022beyond} use clustering to reduce the number of dataset examples on which models need to be trained to obtain high performance.
 Comparatively little work has investigated properties of the resulting clusters. \citet{yangneurons} experiment with combining pairs of CIFAR-10 classes into superclasses. Although extensive training leads to a ``neural collapse'' phenomenon where representations collapse to class centroids~\cite{papyan2020prevalence}, they show that the representations within a given class nonetheless remain structured and it is possible to recover the subclasses via clustering. Our work seeks to study this clusterability property in a systematic way, by intervening on not just the dataset structure, but also the training objective as well as the model architecture.

As discussed in the introduction, our work is related to, but distinct from, work that has proposed specialized deep representation learning methods for clustering. Approaches such as DEC~\cite{xie2016unsupervised}, SCAN~\cite{van2020scan}, SPICE~\cite{niu2022spice}, and TCL~\cite{li2022twin} assign images to clusters after one or more stages of model training using a specialized clustering objective. Subclass distillation~\cite{muller2020subclass} learns subclasses of dataset classes which are then used to improve distillation. Other methods aim to cluster a novel set of images using knowledge gleaned from another image set~\cite{hsu2018learning,han2019learning,Han2020Automatically,zhong2021neighborhood}, a task that has been termed novel category discovery (NCD). Some self-supervised learning methods such as DeepCluster~\cite{caron2018deep} and SwAV~\cite{caron2020unsupervised} also perform clustering, but with the aim of learning good representations for linear classification, semi-supervised learning, and transfer learning, rather than producing cluster assignments for training images. In our work, we experiment with using learned representations to perform clustering on downstream datasets, whose classes overlap with ImageNet categories to different extents.

Clustering is only one of many methods that can be used to simplify or otherwise gain insight into data. Other methods seek to find influential training points for a specific model prediction~\cite{koh2017understanding,ilyas2022datamodels}, training datapoints that are difficult for the model to learn \cite{pmlr-v139-jiang21k,agarwal2022estimating,siddiqui2022metadata}, subsets of training datapoints that are sufficient to achieve high performance~\cite{mirzasoleiman2020coresets,paul2021deep,killamsetty2021grad,sorscher2022beyond}, or types of data on which the model is consistently poor~\cite{yeh2020completeness,singla2021understanding,eyuboglu2022domino,wiles2022discovering,d2022spotlight}. We demonstrate via comparison with other common representation understanding techniques (e.g. linear probe), that subclass clustering offers a distinctive lens through which to study what neural networks can learn from incomplete labels.
\section{Experiment setup}
\paragraph{Datasets}
To obtain datasets with natural subclass-superclass hierarchy, we leverage BREEDS \cite{santurkar2020breeds}, a subset of ImageNet that follows a modified WordNet hierarchy to better suit object recognition tasks. Compared to the standard WordNet hierarchy, BREEDS makes two major changes: (i) BREEDS places nodes under a common ancestor only if they share some visual information, whereas the WordNet hierarchy is sometimes defined in abstract ways (e.g., ``umbrella'' and ``roof'' are both subclasses of ``covering''); and (ii) nodes of similar specificity (e.g., “dog” and “cat”) share the same distance from the root. These changes allow us to study how label hierarchy is reflected in neural network representations in a controlled fashion.

Our experiments involve three benchmarks suggested in the original paper: entity-13, living-17 and nonliving-26. Dataset names are
based on the root of the subtree in WordNet where nodes are taken from, and the resulting number of superclasses. We modify entity-13 so that each superclass contains only 4 subclasses (i.e. original ImageNet classes), resembling the structure of the other two datasets.
Although substantially smaller than the pre-training datasets of top-performing vision models, all BREEDS datasets still contain $>$60,000 images, and they are the largest datasets of which we are aware with a carefully constructed hierarchy.
Refer to Table \ref{tab:breeds_datasets} in the Appendix for more details.

\paragraph{Models}
Most of our experiments involve ResNet-50 \cite{he2016deep} and VGG-16 \cite{simonyan2014very}, two architectures that have been extensively studied in literature and differ in many ways. Section \ref{sec:normalization_layers} further explores one such difference --- the presence of normalization layers. In the architecture ablation study, we also report clustering performance of ViTs \cite{dosovitskiy2020image}. 

\paragraph{Clustering method \& evaluation}
For each BREEDS dataset, we train models using corresponding images from the train split and only the superclasses as the target labels. For evaluation, we extract embeddings of the corresponding validation split for each dataset from a certain layer, and perform clustering \textit{within each superclass}: the number of clusters is set to be the number of subclasses per superclass, multiplied by some overclustering factor between 1 and 10, to account for the presence of more fine-grained structure beyond the original ImageNet labels. 

Previous work indicates that agglomerative clustering is among the best clustering methods for representations obtained from image classifiers~\citep{monath2021scalable}. We provide a comparison between k-means and agglomerative clustering with various linkage criteria in Table \ref{tab:kmeans_vs_agg} in the Appendix. This initial exploration leads us to adopt the agglomerative clustering based on Ward linkage \cite{ward1963hierarchical} in our experiments. This clustering method is scalable to large number of samples and clusters, and unlike k-means, it is deterministic. We find that compared to using raw embeddings, normalizing the embeddings to have unit norms also helps boost clustering performance while preserving the overall trend, and thus adopt this approach in all experiments below.

To assess cluster quality, we use both purity and adjusted mutual information (AMI). We find that the two metrics yield very similar trends across all of our experiments (see Appendix \ref{app:ami_vs_purity}). To compute purity, we label all samples in each cluster with the class that is most frequent in the cluster, and measure the accuracy of this assignment with respect to the true labels. AMI measures mutual information between the predicted clusters and the true labels, adjusted for chance. Both metrics allow permutation of cluster labels, thus accounting for the possibility of each subclass being split up into multiple clusters (e.g. when overclustering factor $>$ 1). They also both range from 0 to 1, with a higher value indicating better alignment between the clusterings and ground-truth labels. Implementations of clustering algorithms and metrics are done through \texttt{scikit-learn}.

In the subsequent sections, we investigate how the clustering performance with respect to the true subclass structure may be impacted by various components of the learning process --- training data, loss function and model architecture --- in settings where the target data comes from the same or different distributions as the training set.

\section{Clusterability trends within a network}
\begin{figure}[]
\begin{minipage}{0.58\textwidth}
\includegraphics[trim=0 0 0 0,clip,width=\linewidth]
{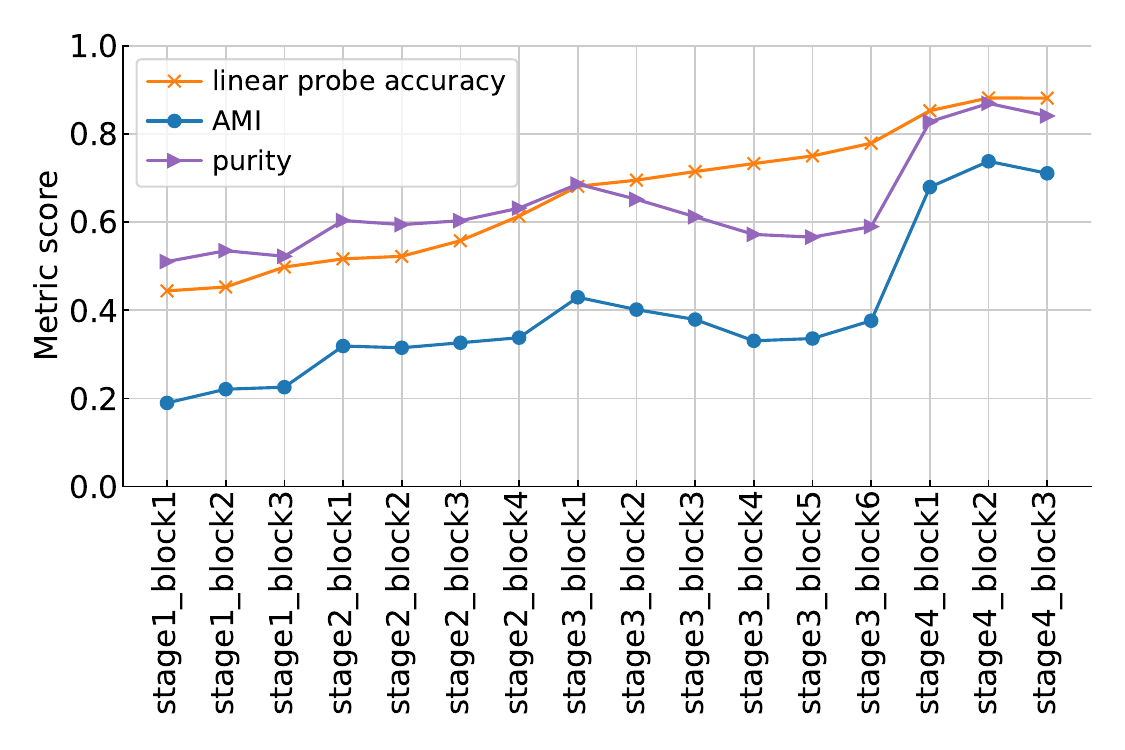}
\end{minipage}\hfill
\begin{minipage}{0.35\textwidth}
\caption{\textbf{Clustering performance is not fully aligned with linear probe accuracy.} We compare AMI and purity of subclass clustering to linear probe performance for different layers in a ResNet-50 trained on entity-13-shuffled. Although all metrics generally increase with depth, clustering evaluations are not perfectly correlated with linear probe accuracy, and reveal greater variations across layers that belong to the same residual block.
}
\label{fig:linear_probe}
\end{minipage}
\vskip -2.2em
\end{figure}

We track the clusterability of subclasses in different layers throughout each trained network. 
Clustering performance, measured with AMI and purity scores, shows an overall increasing trend with depth. This is not surprising as deeper layers have been shown to learn more complex features so as to predict the labeled classes well. A similar phenomenon has been observed with existing tools such as linear probes \cite{alain2017understanding}. For each of the layers from which we obtain embeddings and perform clustering, we also train a linear probe to map from the layer representation to the \textit{subclass} labels. Our clustering metrics and linear probe accuracy both generally increase with depth, but while linear probe accuracy is nearly monotonic, subclass clustering is not (Figure \ref{fig:linear_probe}).

Appendix \ref{app:evolution_clusterability} contains additional results, as well as visualizations of image clusters found at different stages of a ResNet-50. We also fix a layer and analyze clusterability over the course of training for models trained on different BREEDS datasets (Appendix Figure \ref{fig:ami_over_step_vgg}). We find that in most cases, AMI across layers fluctuates substantially throughout training. The only exception is training with entity-13-shuffled (refer to Section \ref{sec:shuffling} for its construction), where AMI stays mostly unchanged for the majority of the network after epoch 100 (i.e. 25\% of the total training time), and only the clusterability of the fully-connected layers continues to rise over time.

\section{The role of training data}
\subsection{Shuffling the dataset hierarchy} \label{sec:shuffling}
\begin{figure}[h]
\vskip -1em
\includegraphics[trim=0 0 0 0,clip,width=0.48\linewidth]
{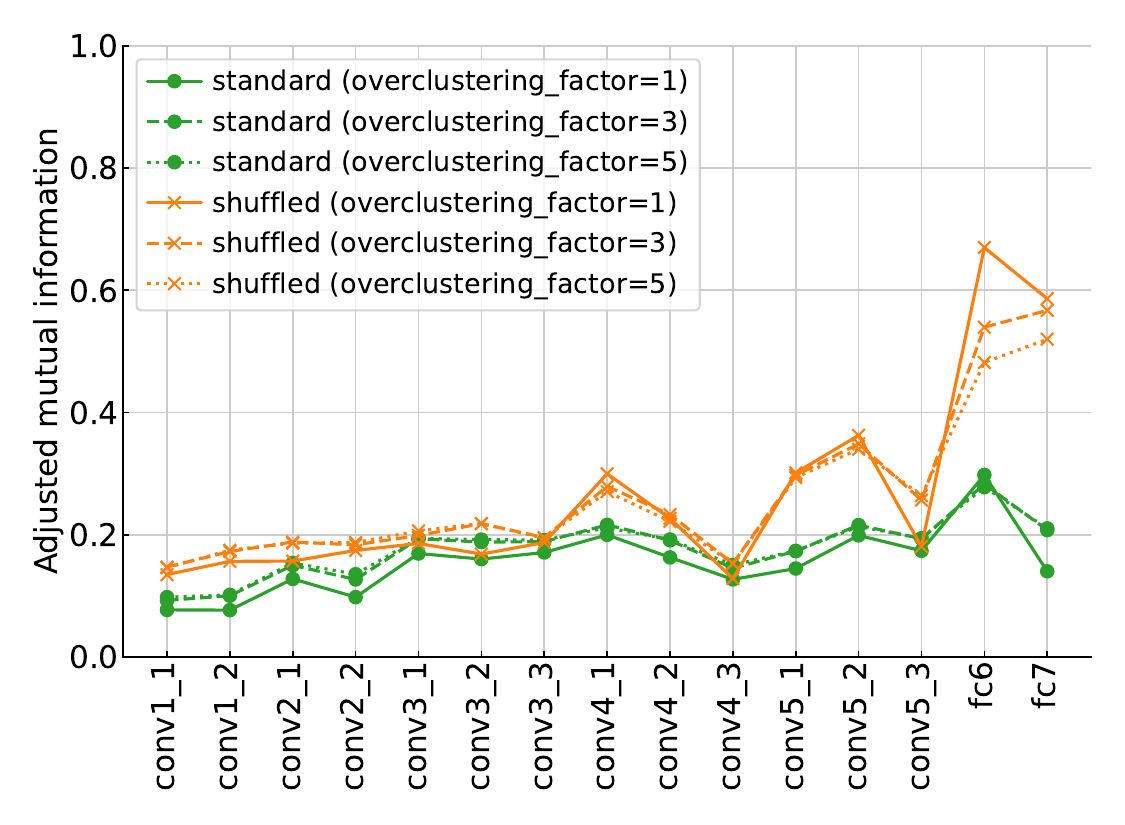}
\includegraphics[trim=0 0 0 0,clip,width=0.48\linewidth]
{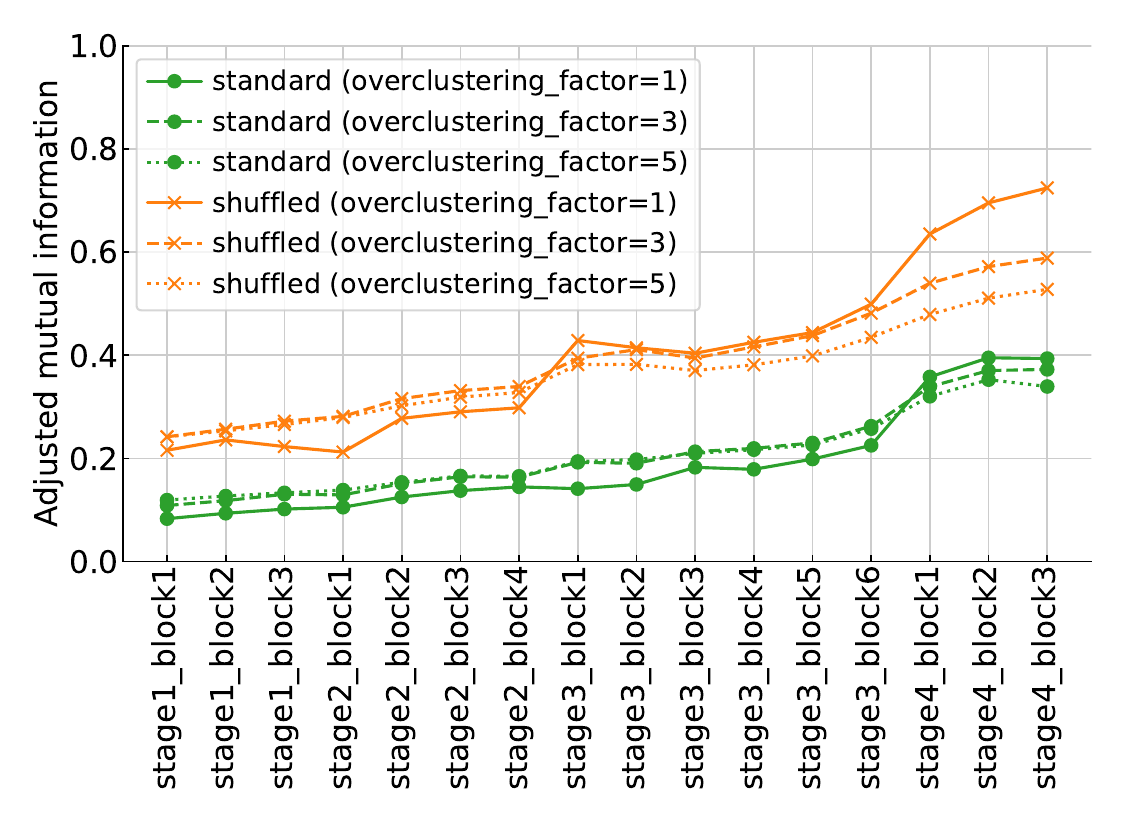}
\vskip -0.5em
\caption{\textbf{Clustering performance improves substantially when labeled classes contain heterogeneous subclasses.} We randomize the subclass-superclass mapping in entity-13 and find that this increases AMI across layers for both VGG-16 (left) and ResNet-50 (right). The improvements in clustering quality also tend to be the largest near the end of either network.
}
\label{fig:shuffle_vs_standard}
\vskip -0.5em
\end{figure}
When trained on superclasses drawn from a hierarchical dataset, a network could pick up only features that are shared across the subclasses of each superclass, or it could instead learn separate features that distinguish each subclass, and then combine these subclasses to predict the superclass. Our clustering evaluation suggests that the strategy that networks use depends on the properties of the training data.

For standard training on BREEDS (green lines in Figure \ref{fig:shuffle_vs_standard}), AMI values remain low but non-zero throughout the network. We posit that this is due to the natural hierarchy present in the training data: although images under each superclass belong to different subclasses, they are still more visually similar to one another than images from other superclasses. As an intervention, we randomly map each subclass to one superclass in entity-13, while still keeping the number of subclasses per superclass constant. We call this variant entity-13-shuffled. As shown in Figure \ref{fig:shuffle_vs_standard}, making the images under each superclass much more heterogeneous greatly improves subclass clusterability (orange lines).
Appendix \ref{app:data} contains additional experiments with the shuffled version of living-17 dataset. This result suggests that when labeled classes contain heterogeneous subclasses without a lot of visual features in common, the “simplest” strategy for the model to predict a labeled class may be to learn a different set of features for each subclass. This in turn leads to high subclass clusterability. 

\vspace{0.25em}
\subsection{Impact of training distribution on clustering performance for external datasets} \label{sec:data_ood_perf}
\vspace{0.25em}
\begin{figure*}
\vskip -1.5em
\includegraphics[trim=0 0 0 0,clip,width=1.01\linewidth]
{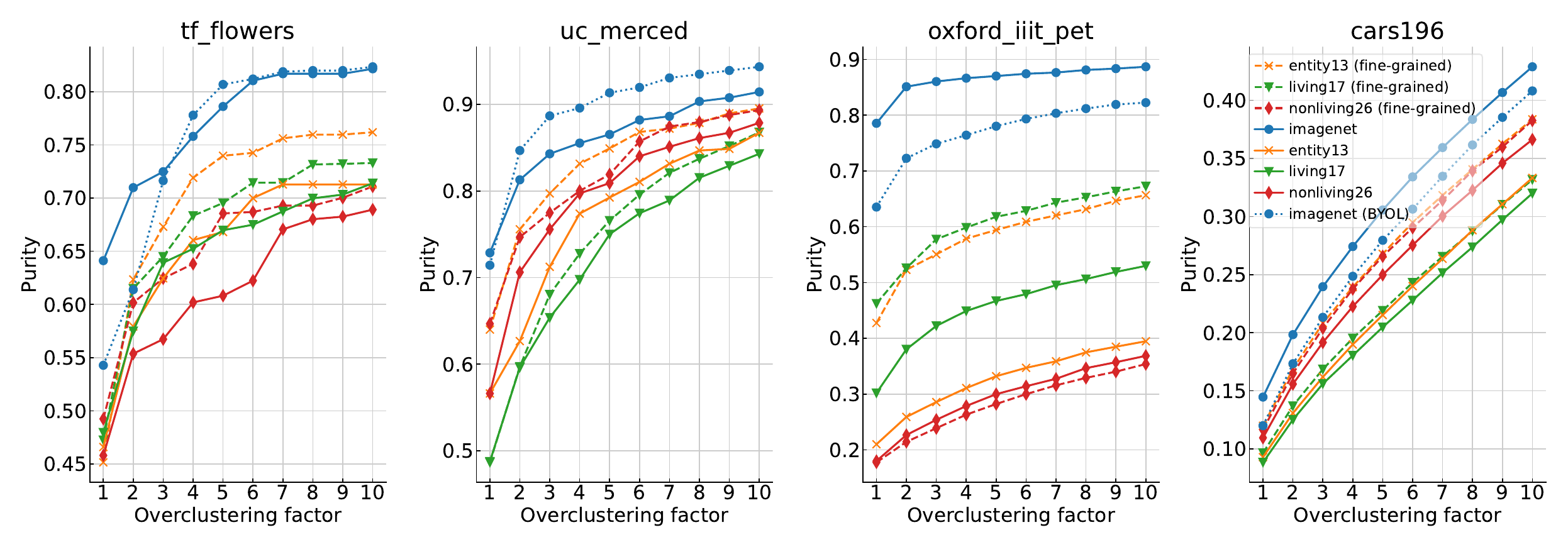}
\vskip -0.5em
\caption{\textbf{When using pre-trained embedddings to recover classes in external datasets, training with subclass labels yields only marginally better clusterability than training with superclass labels.} We evaluate the ability of ResNet-50 models, trained on ImageNet or different BREEDS datasets, to recover the original classes in four downstream datasets. We cluster the image embeddings of all examples in each dataset and measure how well the cluster assignments align with ground-truth labels. 
Using fine-grained (i.e. subclass) labels during pre-training offers some improvement over using superclass labels, but the improvement tends to be relatively small, suggesting that subclass information provides little benefit when detecting class structure in out-of-distribution data. We also compare with the self-supervised model BYOL \cite{grill2020bootstrap}, which, despite not having access to any ground-truth labels during training, offers good downstream clustering performance, especially on coarse-grained data.
}
\label{fig:ood_IN_subsets}
\vskip -2.0em
\end{figure*}
Having examined how hierarchical structure in the training data may affect clustering performance on the \textit{same} data distribution, we next explore the implication of varying image distribution and label granularity on the ability to recover class structure in \textit{external} datasets. We use four downstream datasets with different degrees of overlap with ImageNet and granular structures:
\begin{itemize}[leftmargin=6pt,topsep=0pt,itemsep=0pt,parsep=4pt]
\item Tf\_flowers \cite{tfflowers}: an image dataset of 5 classes of flowers, with `daisy' also being an ImageNet class.
\item UC Merced \cite{Nilsback08}: a 21-class land use remote sensing image dataset. Only the `buildings' class may potentially overlap with ImageNet training data.
\item Oxford-IIIT Pets \cite{parkhi12a}: a 37-category image dataset of dog and cat breeds. Out of the four in the list, this dataset has the most overlap with ImageNet, sharing 25 of its classes.
\item Cars196 \cite{KrauseStarkDengFei-Fei_3DRR2013}: a dataset of 196 classes of cars based on make, model and year. This is the most fine-grained dataset in our evaluation; ImageNet contains cars, but only use general type labels (e.g. `sports car').
\end{itemize}

For each test set, we extract the penultimate layer embeddings from a ResNet-50 trained on BREEDS data, and cluster the embeddings into $N$ clusters, with $N$ being the original number of classes in the test set multiplied by some overclustering factor between 1 and 10. 
In Figure \ref{fig:ood_IN_subsets}, we find that except for Cars196, training on ImageNet or just subsets of ImageNet allows us to recover class structure in external datasets to a large extent, even if these datasets have little in common with ImageNet. 

Although using subclass labels during pre-training generally improves cluster purity in downstream datasets compared to using superclass labels, this improvement tends to be relatively small (Figure \ref{fig:ood_IN_subsets}). Consequently, the majority of the signals for clustering out-of-distribution data comes from superclass supervision. The only exception to this observation is Oxford-IIIT Pets, which also involves fine-grained classification and whose classes directly overlap with the subclasses in entity-13 and living-17 (see Table \ref{tab:breeds_datasets}). For this task, we observe significant increases in cluster purity when using representations of models trained with subclass supervision.
\section{The role of loss function}
\vspace{0.25em}
\subsection{Supervised versus Self-supervised training}
\vspace{0.25em}
In Figure \ref{fig:ood_IN_subsets}, we also compare clustering performance out-of-distribution between a ResNet-50 trained on the full ImageNet in a supervised manner, versus a ResNet-50 trained in a self-supervised manner without any ground-truth labels. For the latter, we use Bootstrap Your Own Latent (BYOL) and evaluate the model checkpoint open-sourced by \citet{grill2020bootstrap}.

We find that BYOL offers higher cluster purity compared to using label supervision with less data, even for downstream datasets whose classes directly overlap with the subclasses in the BREEDS training set, e.g. Oxford-IIIT pet and living-17 share dog and cat classes. However, given the same training set, supervised training produces better clustering performance than self-supervised training in fine-grained classification tasks (i.e. Oxford-IIIT pet and Cars196, see Figure \ref{fig:ood_IN_subsets}).

\vspace{0.25em}
\subsection{Squared error versus Softmax cross-entropy} \label{sec:squared_vs_softmax}
\begin{figure}[]
\begin{minipage}{0.57\textwidth}
\includegraphics[trim=0 0 0 0,clip,width=\linewidth]
{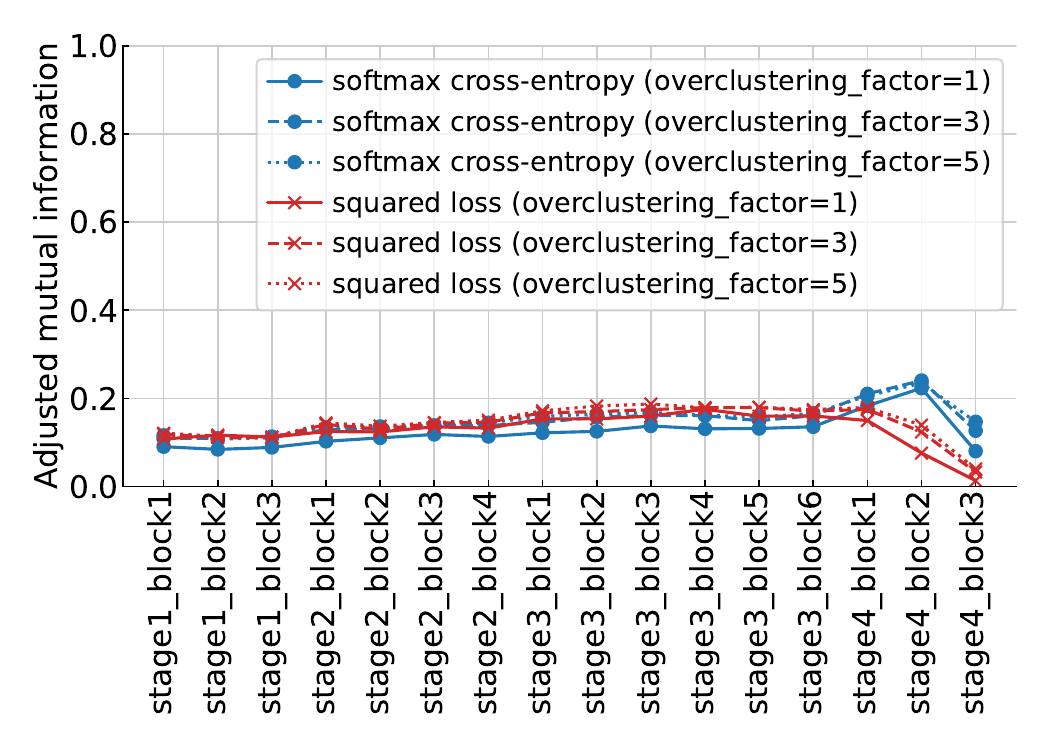}
\end{minipage}\hfill
\begin{minipage}{0.4\textwidth}
\caption{\textbf{Training with squared error yields the same clusterability as softmax cross-entropy, except at the later stages of the network}. We compare AMI across layers for training a ResNet-50 with squared error versus softmax cross-entropy loss on living-17 dataset. The two objective functions produce similar AMI profiles for most of the network, except at the later layers where squared error sees a drop in AMI, possibly due to overfitting to the training classes \cite{kornblith2021better}.
}
\label{fig:squared_vs_softmax}
\end{minipage}
\vskip -2.5em
\end{figure}

\vspace{0.25em}
We next zoom in on the supervised setting and explore variations in loss functions for models trained in a supervised manner. First, we experiment with using squared loss instead of the standard softmax cross-entropy objective. Figure \ref{fig:squared_vs_softmax} shows the AMI across layers of a ResNet-50 trained with either objective. We find that squared error produces similar clustering performance as softmax, except for at the last few layers of the network where squared loss sees a drop in AMI. This observation is in line with previous results showing that squared loss produces greater class separation and more overfitting to the training classes compared to softmax \cite{kornblith2021better}. 

\begin{figure}[]
\begin{minipage}{0.57\textwidth}
\vskip -1.5em
\includegraphics[trim=0 0 0 0,clip,width=\linewidth]
{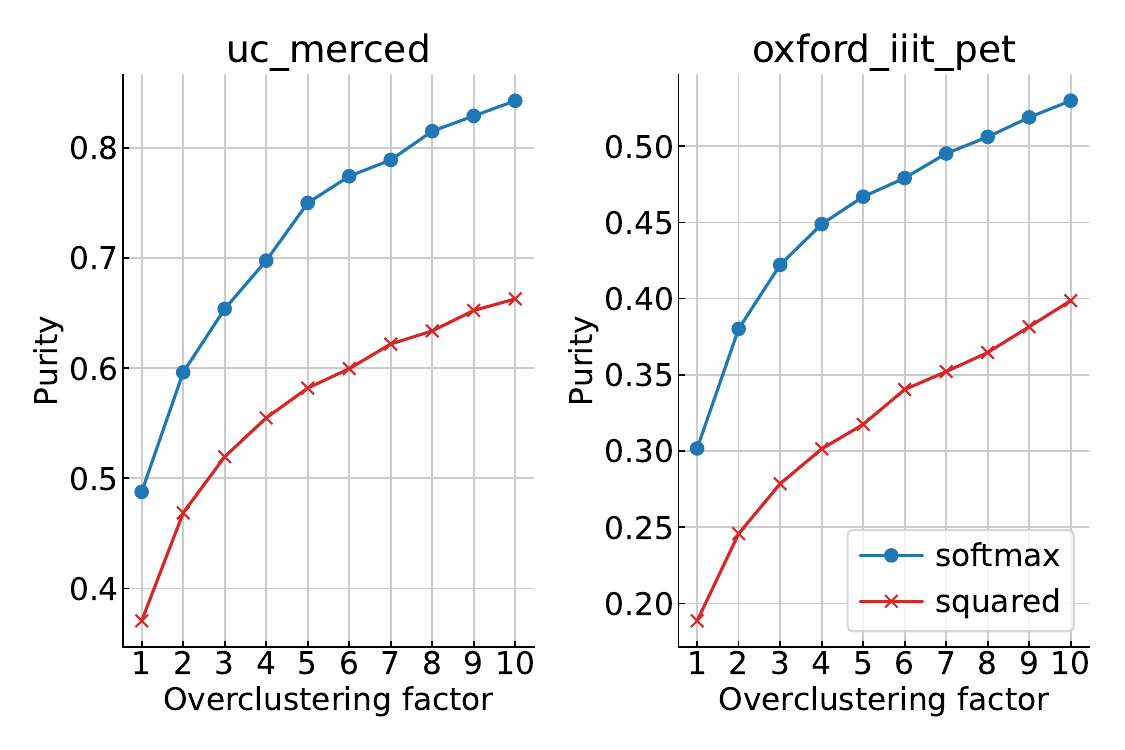}
\end{minipage}\hfill
\begin{minipage}{0.4\textwidth}
\caption{\textbf{Compared to softmax cross-entropy, embeddings trained with squared loss yield lower clustering performance on external datasets.} Training a ResNet-50 on living-17 dataset with squared loss also performs worse than training with softmax cross-entropy at recovering the class structure in out-of-distribution data. The full evaluations on all four external datasets can be found in Appendix Figure \ref{fig:ood_squared_vs_softmax_full}.
}
\label{fig:ood_squared_vs_softmax}
\end{minipage}
\vskip -1.5em
\end{figure}

When we use the representations obtained from these two loss functions to recover the class structures in external datasets (see the list in Section \ref{sec:data_ood_perf}), we find that squared error also leads to significantly lower clusterability out-of-distribution, similar to what \citet{kornblith2021better} observe with transfer performance. We show the clustering results on UC Merced and Oxford-IIIT Pet datasets in Figure \ref{fig:ood_squared_vs_softmax} and defer results on the other two external datasets to Figure \ref{fig:ood_squared_vs_softmax_full} in the Appendix. We repeat the same experiment with other BREEDS datasets; see Appendix \ref{app:loss_functions} for more details.

\vspace{0.25em}
\subsection{Impact of loss functions on clustering performance out-of-distribution}
\begin{figure*}
\includegraphics[trim=0 0 0 0,clip,width=1.01\linewidth]
{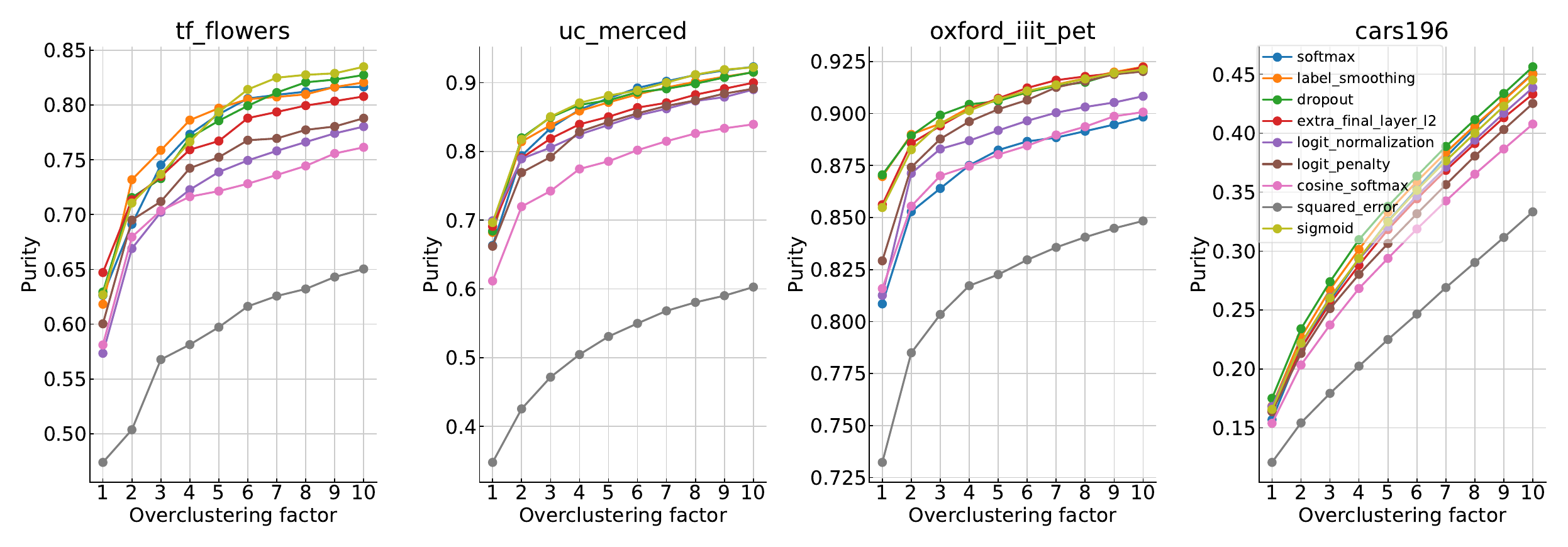}
\vskip -0.5em
\caption{\textbf{Training objectives affect the clustering performance on external data; although softmax cross-entropy has been shown to yield better linear transfer than other losses, it actually produces worse clusterability.} We take the models open-sourced by \citet{kornblith2021better} that have been trained on the full ImageNet with nine different loss functions, and evaluate how well the embeddings of these models can recover classes in external datasets. Across four evaluation sets, squared loss yields the worst clusterings, followed by cosine softmax loss. While \citet{kornblith2021better} find that softmax offers the highest accuracies for linear transfer and $k$-nearest neighbors on external tasks, our evaluation metric shows that in the case of subclass clustering, other loss functions and regularization methods such as sigmoid and dropout may be more ideal.
}
\label{fig:ood_loss_functions}
\vskip -1em
\end{figure*}

\vspace{0.25em}
To extend the findings from the previous section and investigate a wider range of training objectives, we evaluate open-sourced models from \cite{kornblith2021better} trained on the full ImageNet dataset with nine different types of losses and regularization methods. While \citet{kornblith2021better} focus on the transferability of the resulting hidden representations, our work measures the quality of the representations by the ability to recover class structures in external datasets (see the dataset list in Section \ref{sec:data_ood_perf}). Clustering results averaged across three seeds are reported in Figure \ref{fig:ood_loss_functions}.

We again find that the clustering performance from using models trained with squared error is significantly worse than from models trained with other objective functions. However, although \citet{kornblith2021better} find that unregularized softmax cross-entropy provides the most transferable features --- measured by the accuracies of linear transfer and $k$-nearest neighbors on a variety of tasks (including Oxford-IIIT Pets)
--- our clustering evaluation shows that softmax can yield inferior clusters compared to other loss functions
, especially on the Oxford-IIIT Pets classification task. More specifically, when training a fixed model with the goal of using it to uncover structure in downstream related datasets, regularizers such as dropout and label smoothing or alternative losses such as sigmoid cross-entropy can be very useful, even though they result in worse linear transfer.

\section{The role of model architecture}
\subsection{VGG versus ResNet} \label{sec:vgg_vs_resnet}
Having examined the influence of training set and loss function on the clustering of the hidden representations, we next look at another widely studied component of training: model architecture.
We track the clusterability of subclasses in a range of layers making up each architecture. In ResNet-50, we examine only the layers following a residual connection. In VGG-16, we look at all convolutional layers as well as two fully connected layers preceding the output layer. 

On entity-13 and entity-13-shuffled, we observe that the first fully connected layer (out of three) in VGG-16 is where subclass clustering is most aligned with ground truth. In contrast, clusterability tends to increase throughout and peak at the end for ResNet-50 architecture (Figure \ref{fig:bn_vs_ln}). Refer to Appendix \ref{app:model_architecture} for corresponding results on nonliving-26 dataset.

\vspace{0.25em}
\subsection{Normalization layers}
\label{sec:normalization_layers}
\vspace{0.25em}
We seek to better understand the differences in clustering behavior between VGG-16 and ResNet-50  by looking at the principal differences in the building blocks of the two architectures. Besides the absence of residual connections, VGG-16 also has no batch normalization layer \cite{ioffe2015batch}. Regularization benefits of batch norm have been studied extensively in prior work \cite{santurkar2018does, morcos2018importance, luo2018towards}.

We proceed to add batch normalization layers after every convolutional and dense layer in the original VGG-16 architecture. We also experiment with layer normalization \cite{ba2016layer}, another common normalization technique. Clustering results for entity-13 and entity-13-shuffled datasets are shown in Figure \ref{fig:bn_vs_ln} (orange lines). Superclass classification accuracies can be found in Appendix Table \ref{tab:acc}.

We find that normalization layers boost clustering performance only for the intermediate layers of VGG-16, with a larger boost when superclasses contain heterogeneous subclasses. In addition, batch normalization appears to yield larger improvements than layer normalization. While prior work has suggested that batch normalization can reduce class selectivity of individual features and encourage generalization \cite{morcos2018importance}, we find that averaged across three random seeds, the maximum AMI attainable across layers of VGG-16 does not change with batch normalization, only shifts to an earlier layer. 
\begin{figure}[]
\centering
\includegraphics[trim=0 0 0 0,clip,width=0.49\linewidth]
{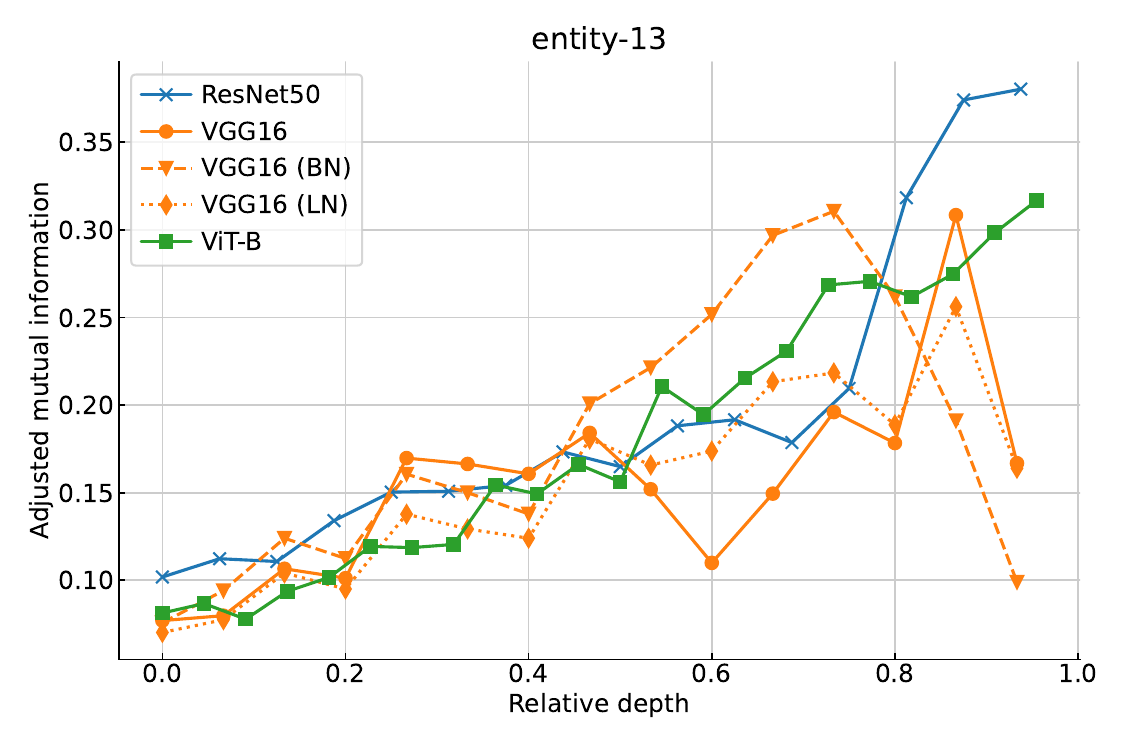}
\includegraphics[trim=0 0 0 0,clip,width=0.49\linewidth]
{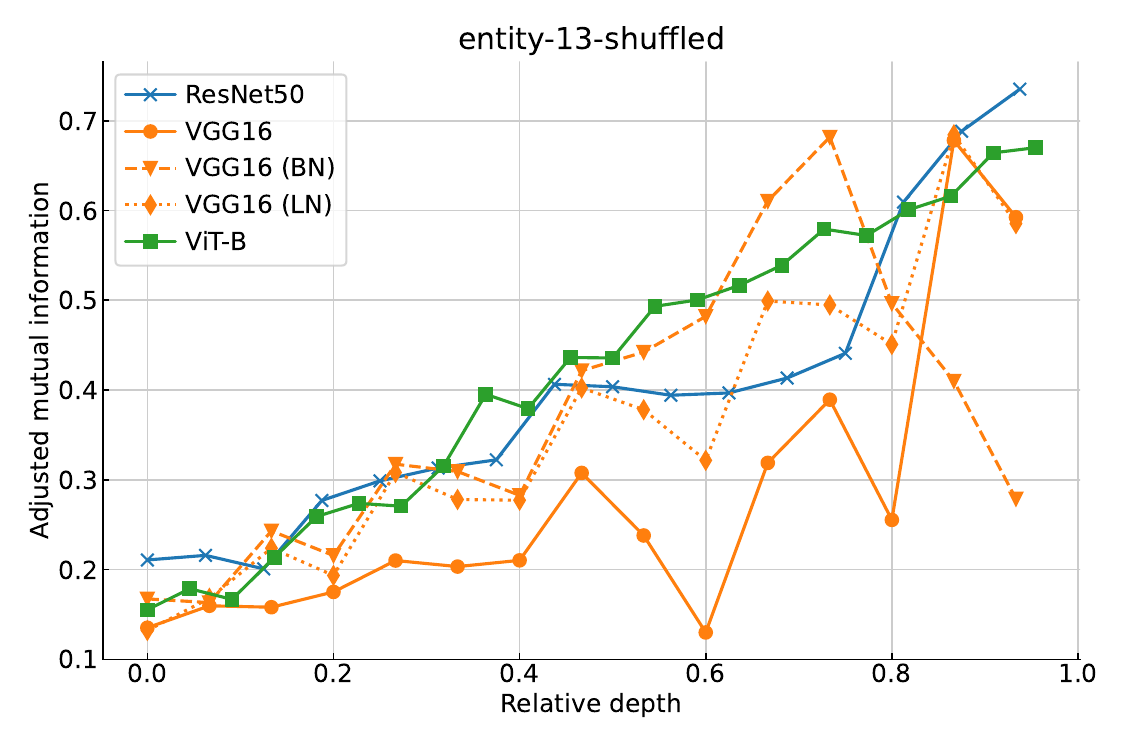}
\vskip -0.5em
\caption{\textbf{Both the model architecture and the type of normalization affect which layer provides the best clustering.} We compare how clustering performance changes with depth, for different model architectures trained on entity-13 (left) and entity-13-shuffled (right), averaged across three training runs. We find that both the architecture family and the normalization method affect where subclass clusterability peaks. Furthermore, layer normalization and batch normalization yield different clustering behaviors, but the maximum AMI attainable from a VGG-16 does not increase with normalization. Notably, ViT-B can yield worse clusterability compared to ResNet-50 and VGG-16.}
\label{fig:bn_vs_ln}
\vskip -1.5em
\end{figure}

\vspace{0.25em}
\subsection{Vision Transformers} \label{sec:vit}
\vspace{0.25em}
We also experiment with training ViT-B models \cite{dosovitskiy2020image} on entity-13 and entity-13-shuffled, measuring clusterability at the residual connections that follow self-attention and MLP layers in each encoder block. The clustering performance across layers could be found in Figure \ref{fig:bn_vs_ln} (green line). We find 
that the maximum AMI attained by ViT-B is actually lower than that of ResNet-50, and is worse than VGG-16 in the case of entity-13-shuffled. 

\section{Comparing clusterings across networks} \label{sec:across_networks}
\begin{figure*}[]
\vskip -1em
\subfloat{
    \includegraphics[trim=0 0 0 0,clip,width=0.49\linewidth]{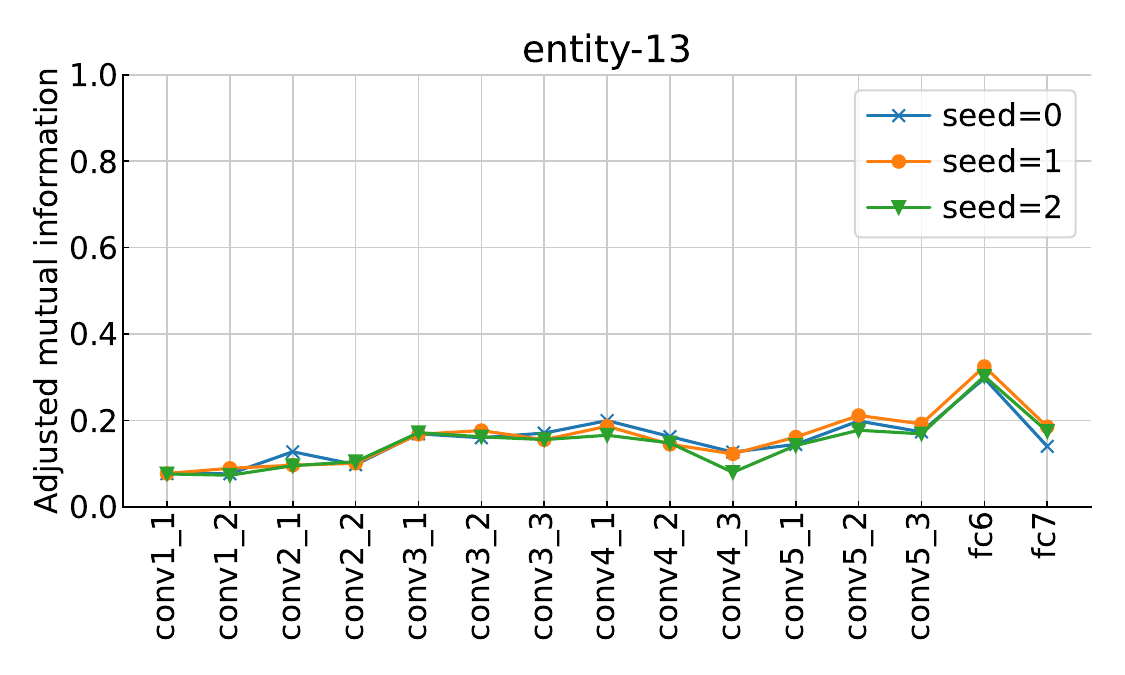}
}\hfill
\subfloat{
  \includegraphics[trim=0 0 0 0,clip,width=0.49\linewidth]{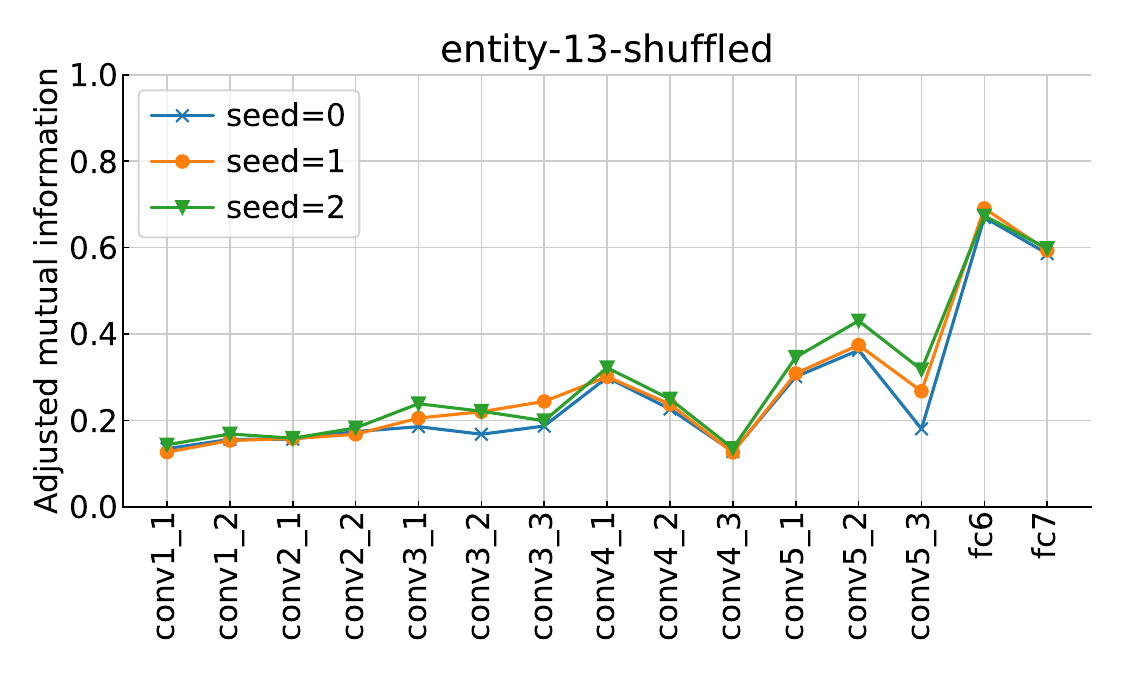}
}
\\ \vskip -2.5em
\subfloat{
  \includegraphics[trim=0 0 0 0,clip,width=0.49\linewidth]{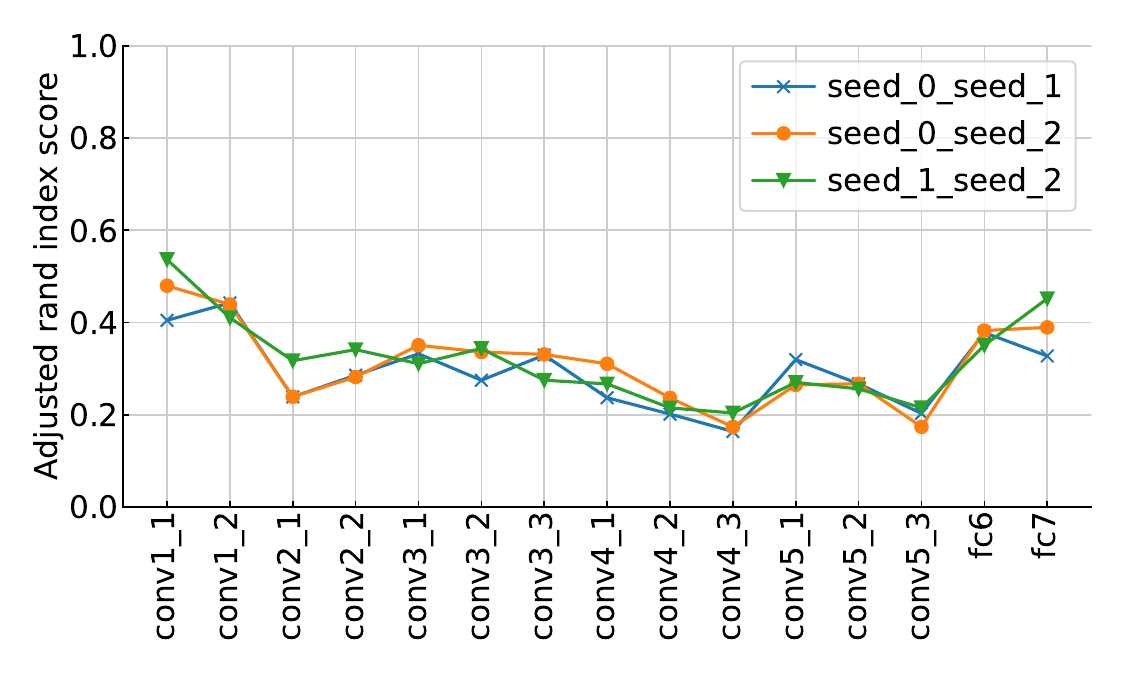}
}\hfill
\subfloat{
  \includegraphics[trim=0 0 0 0,clip,width=0.49\linewidth]{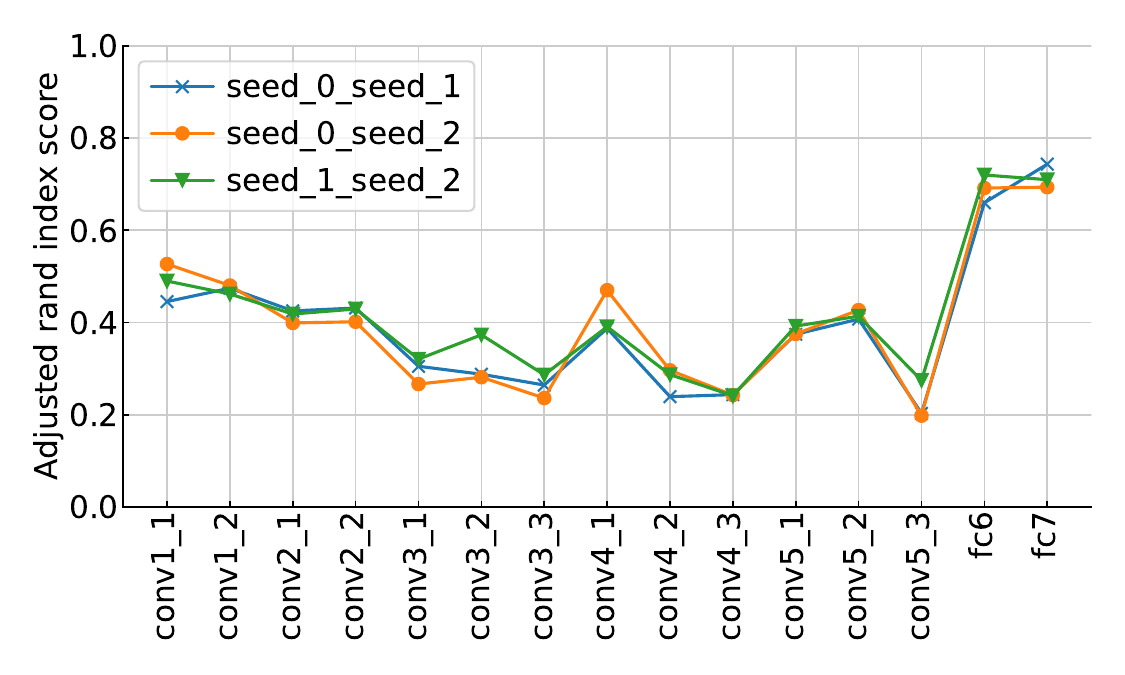}
}
\vskip -0.5em
\caption{\textbf{Different training runs yield similar layer-wise clustering performance, but cluster assignments are highly inconsistent across runs.} The top row shows the AMI across layers for three training runs of VGG-16 on entity-13 (left) and entity-13-shuffled (right). These random seeds produce highly similar AMI profiles. However, when measuring how aligned the actual clusters are across runs with adjusted Rand index, we find that the ARI scores are relatively low for most layers (bottom row) --- the images assigned to each cluster vary substantially across runs. An exception is high cluster consistency found in the last few layers of entity-13-shuffled model, where we also observe a corresponding rise in AMI. The same trends apply to ResNet-50; see Appendix Figure \ref{fig:ARI_across_seeds_resnet50}.
}
\label{fig:ARI_across_seeds_vgg16}
\vskip -2em
\end{figure*}
The previous sections examine how factors such as dataset structure, model architecture and training objective may affect the ability to recover subclasses from the hidden representations of a single model. We next turn to measuring the consistency of the clusters uncovered \textit{across} multiple training runs. In the top row of Figure \ref{fig:ARI_across_seeds_vgg16}, we show the AMI across layers for three random seeds of a VGG-16, trained with the entity-13 (left) and entity-13-shuffled (right) datasets. We observe that different training runs of the same model yield highly similar AMI profiles. Does this mean that these runs produce similar clusters?

To measure similarity of clusterings, we use adjusted Rand index (ARI) \cite{hubert1985comparing}. ARI counts the proportion of similarly classified pairs of data points, ignoring permutations, and corrects for chance. 
An ARI value of 0 indicates that cluster labels for two clusterings are independent from each other, and a value of 1 means the clusterings are identical. 
With this metric, we find that cluster consistency across different training runs is surprisingly low. For VGG-16 trained on standard BREEDS datasets, the similarity of cluster assignments tends to peak at the first convolutional layer of the network. In contrast, the similarity is relatively high both at the start and near the end of a ResNet-50. Refer to Appendix \ref{app:across_seeds} for experiments on other datasets and ResNet-50, as well as visualizations of the clusters found across different training runs for the same superclass.

Figure \ref{fig:ARI_across_seeds_vgg16} also demonstrates that randomizing subclass-superclass mapping helps boost cluster similarity across training runs (bottom right plot), especially for the later layers in the networks. We find that the cluster similarity score is the highest at the same layers that also see high alignment with ground-truth labels (AMI), e.g. \textit{fc6} in VGG-16. Figure \ref{fig:ARI_across_seeds_resnet50} (Appendix) shows the corresponding results for ResNet-50. We also provide a comparison of clustering alignment between every pair of layers found in two different random seeds; see Appendix \ref{app:across_seeds} for the full heatmaps.


\section{Conclusion}
Inspired by previous work that leverages clustering in representation spaces, we perform a systematic study of how standard training components~---~training dataset, model architecture, and loss function~---~affect the subclass clusterability at different layers of the resulting network. 
In the process, we uncover several distinct phenomena, demonstrating the potential of subclass clustering as an useful addition to the toolbox of existing techniques (e.g. linear probes, representation similarity measures) that seek to understand neural network internals. 

In particular, we find that given only superclass information as labels during training, the corresponding subclasses are better separated in the internal representations when they are very different from each other than when they follow a natural hierarchy. Providing supervision at a finer level of this natural hierarchy by training on subclass labels improves the ability to recover class structure in external datasets, but only when these datasets have high domain overlap with the pre-training data. A second major theme of our findings is that linear transferability and clusterability sometimes fail to align, both when varying the depth at which representations are taken and when varying the loss function with which networks are trained.

Our work presents several opportunities for further investigation. One question is whether other clustering techniques can significantly outperform agglomerative clustering. Different methods are likely to yield correlated results, 
but novel ways of applying clustering could provide other kinds of insights into the hidden representations.
Besides, several of our observations are in need of further explanation. In particular, our finding that different normalization techniques lead to different patterns of clustering across layers is somewhat perplexing, and we are unaware of any existing work that has characterized a related phenomenon. Last but not least, future work could explore the possibility of learning a linear transformation on the hidden representations to improve subclass clusterings.

\section*{Acknowledgements}
We thank Pieter-Jan Kindermans, Mitchell Wortsman and Dhruba Ghosh for providing feedback on the manuscript.

\bibliographystyle{abbrvnat}
\bibliography{neurips_2023}
\newpage
\appendix
\counterwithin{figure}{section}
\counterwithin{table}{section}

\section{Training details}
\paragraph{Hyperparameters} Except for variations in loss function, model architecture, and dataset, all of VGG and ResNet models from our experiments are trained with batch size 256 for 400 epochs, using SGD with momentum 0.9 and weight decay 0.0001. ViT-B uses AdamW optimizer and weight decay 0.1. Model weights are updated with exponential moving average decay rate of 0.99. When network architecture involves batch normalization layers, we also set batch norm decay rate to be 0.99. For better training stability, we use learning rate 0.01 for VGGs, 0.001 for ViT-Bs, and 0.1 for ResNets. In experiments involving squared loss, we reduce the learning rate to 0.001. We also employ a cosine learning rate schedule, with 5 warmup epochs.

\begin{table}[h]
\begin{center}
\begin{tabular}{ | m{6em} | m{1.5cm}| m{1.5cm} | m{1.5cm} |  m{15em} | } 
  \hline
  Dataset & Root node & Number of superclasses & Number of subclasses per superclass & Superclass names \\
  \hline
  Living-17 & `living thing' & 17 & 4 & `salamander', `turtle', `lizard', `snake, serpent, ophidian', `spider', `grouse', `parrot', `crab', `dog, domestic dog, Canis familiaris', `wolf', `fox', `domestic cat, house cat, Felis domesticus, Felis catus', `bear', `beetle', `butterfly', `ape', `monkey'\\ 
  \hline
  Nonliving-26 & `non-living thing' & 26 & 4 & `bag', `ball', `boat', `body armor, body armour, suit of armor, suit of armour, coat of mail, cataphract', `bottle', `bus, autobus, coach, charabanc, double-decker, jitney, motorbus, motorcoach, omnibus, passenger vehicle', `car, auto, automobile, machine, motorcar', `chair', `coat', `digital computer', `dwelling, home, domicile, abode, habitation, dwelling house', `fence, fencing', `hat, chapeau, lid', `keyboard instrument', `mercantile establishment, retail store, sales outlet, outlet', `outbuilding', `percussion instrument, percussive instrument', `pot', `roof', `ship', `skirt', `stringed instrument', `timepiece, timekeeper, horologe', `truck, motortruck', `wind instrument, wind', `squash'\\ 
  \hline
  Entity-13 & `entity' & 13 & 4 & `garment', `bird', `reptile, reptilian', `arthropod', `mammal, mammalian', `accessory, accoutrement, accouterment', `craft', `equipment', `furniture, piece of furniture, article of furniture', `instrument', `man-made structure, construction', `wheeled vehicle', `produce, green goods, green groceries, garden truck' \\ 
  \hline
\end{tabular}
\vskip 1em
\caption{Details about the BREEDS datasets used in our experiments. We modify entity-13 dataset compared to what was proposed in the original paper \cite{santurkar2020breeds}, by reducing the number of subclasses per superclass to 4, to match the structure of the other two benchmarks. Note that, since most ImageNet classes contain 1300 training images, the size of each training set is approximately $1300 \times \texttt{num\_superclasses} \times \texttt{num\_subclasses\_per\_superclass}$.}
\label{tab:breeds_datasets}
\end{center}
\end{table}

\section{K-means versus Hierarchical clustering}
Table \ref{tab:kmeans_vs_agg} shows results of our preliminary investigation with several choices of clustering algorithm, applied to the embeddings from the penultimate layer of BREEDS-trained models. Based on purity values obtained, we decide to go with agglomerative clustering using `ward' linkage criterion for our subsequent experiments.

\begin{table}[h]
\begin{center}
\begin{tabular}{ | m{12em} | m{7em}| m{7em} | m{9em}| } 
  \hline
  Method & Purity (entity-13) &  Purity (living-17) & Purity (nonliving-26) \\
  \hline
  K-means & 0.368 & 0.360 &  0.371\\ 
  \hline
  Agglomerative clustering, linkage = `ward' & 0.368 & 0.364 & 0.377\\ 
  \hline
  Agglomerative clustering, linkage = `average' & 0.324 & 0.324 & 0.303 \\
  \hline
  Agglomerative clustering, linkage = `complete' & 0.337 & 0.340 & 0.344 \\
  \hline
  Agglomerative clustering, linkage = `single' & 0.268 & 0.269 & 0.266 \\
  \hline
\end{tabular}
\vskip 1em
\caption{Comparison of cluster purity produced by different clustering methods. We use raw image embeddings from the penultimate layer of a VGG-16 trained on entity-13, living-17 and nonliving-26 datasets.}
\label{tab:kmeans_vs_agg}
\end{center}
\end{table}

\FloatBarrier
\section{Adjusted mutual information versus Purity}\label{app:ami_vs_purity}
\begin{figure}[h!]
\centering
\includegraphics[trim=0 0 0 0,clip,width=0.325\linewidth]
{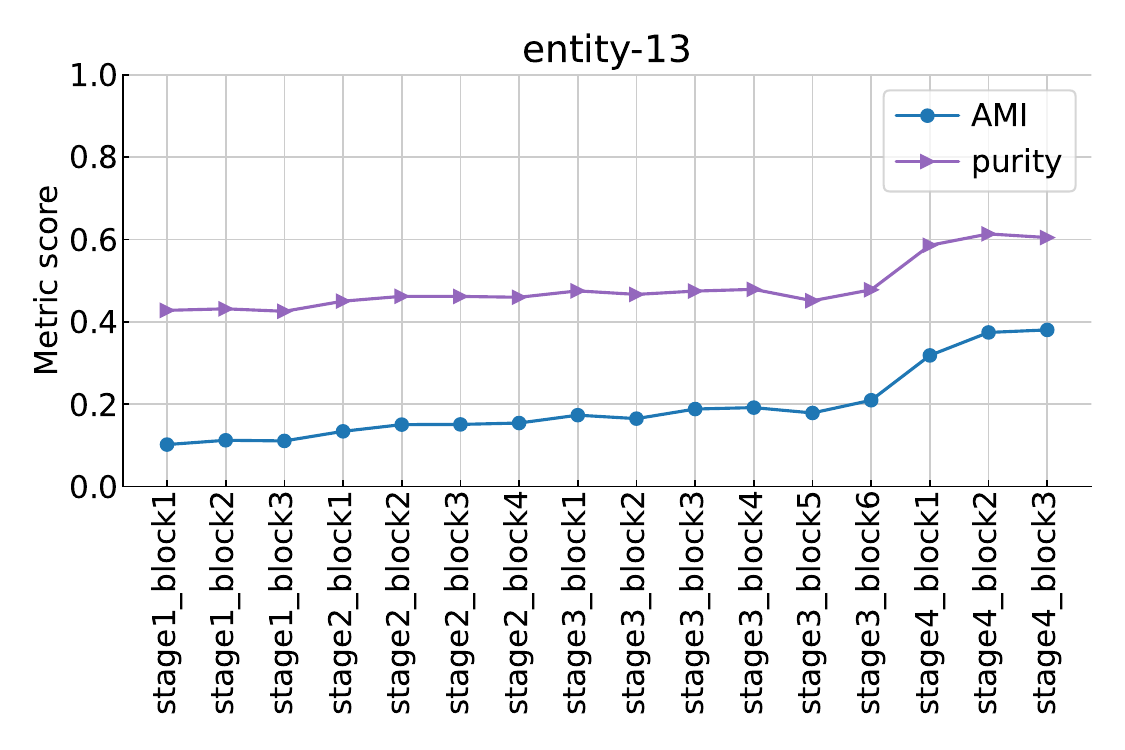}
\includegraphics[trim=0 0 0 0,clip,width=0.325\linewidth]
{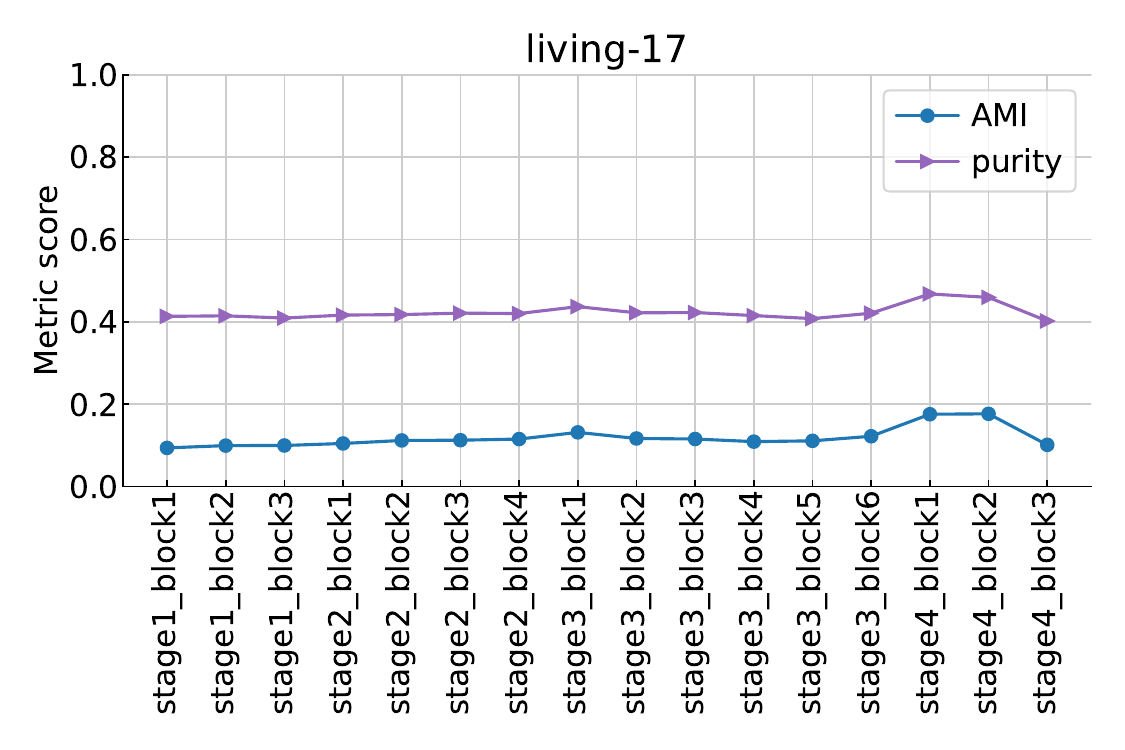}
\includegraphics[trim=0 0 0 0,clip,width=0.325\linewidth]
{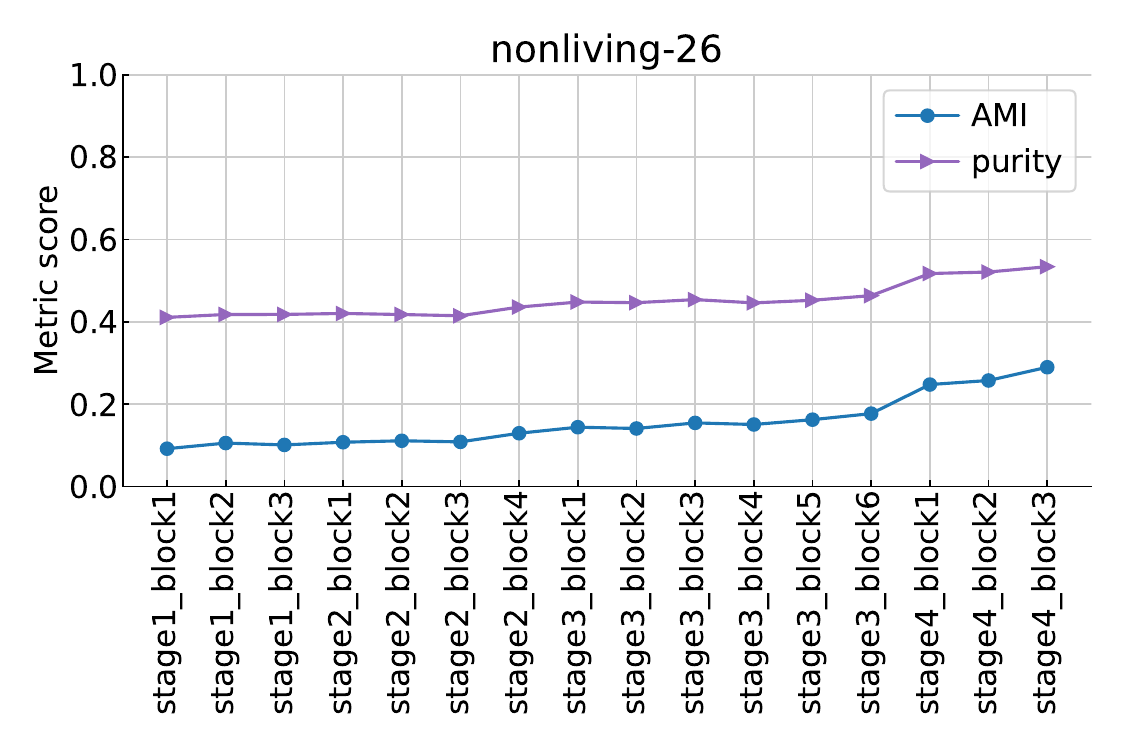}
\caption{We observe similar patterns of clustering performance across layers when using either AMI or purity to measure cluster quality. Here we show results averaged across 3 random seeds for ResNet-50s trained on entity-13 (left), living-17 (middle) and nonliving-26 (right).
}
\label{fig:ami_vs_purity_all_datasets}
\end{figure}

For all BREEDS datasets and models that we experiment with, AMI and purity exhibit highly similar trends across layers --- see Figure \ref{fig:ami_vs_purity_all_datasets} for example.

\FloatBarrier
\section{Evolution of clusterability for a single network} \label{app:evolution_clusterability}
\vspace{1em}
Below we show sample images from each of the 4 clusters uncovered under the superclasses `boat' and `ball' of nonliving-26, using embeddings from different layers of a ResNet50 trained on the same dataset. Each row represents one cluster, and each image visualization is labeled with the ground-truth subclass name:

\begin{figure}[h]
\centering  
\includegraphics[width=\linewidth]{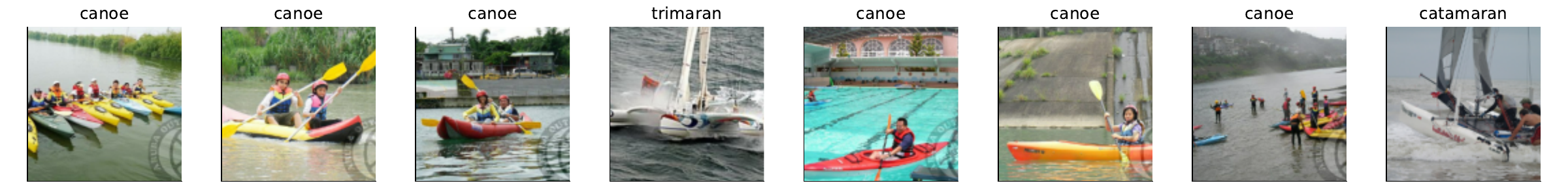}
\includegraphics[width=\linewidth]{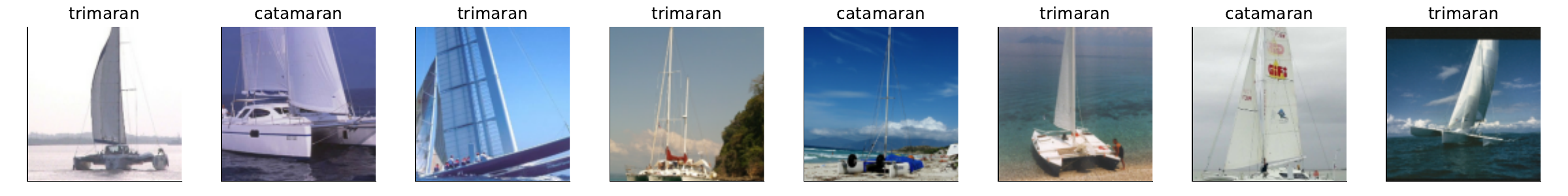}
\includegraphics[width=\linewidth]{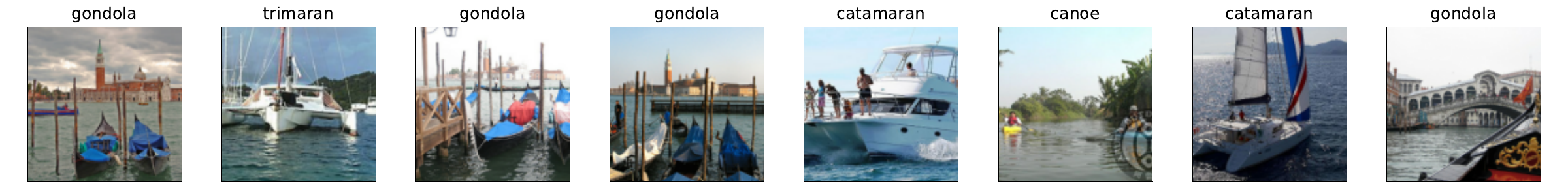}
\includegraphics[width=\linewidth]{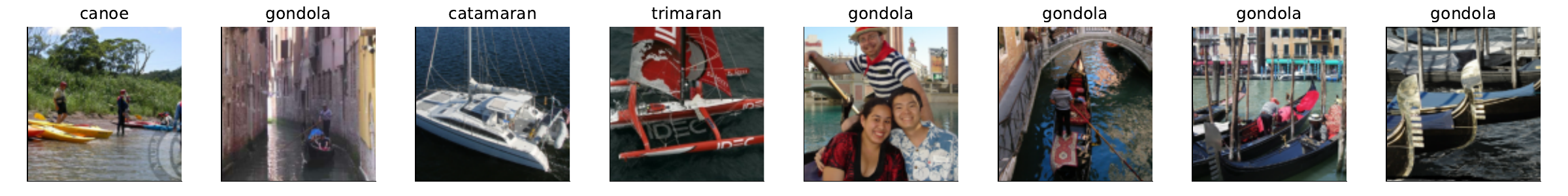}
\vskip -1em
\caption{Sample images from 4 clusters under `boat' uncovered in stage1\_block1 of a ResNet-50.}
\end{figure}

\begin{figure}
\centering
    \includegraphics[width=\linewidth]{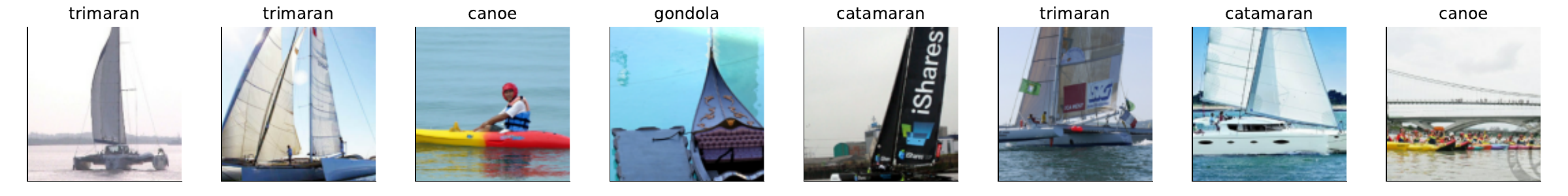}
    \includegraphics[width=\linewidth]{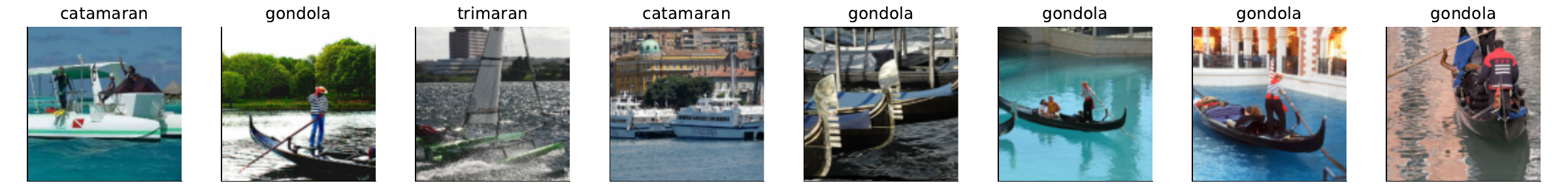}
    \includegraphics[width=\linewidth]{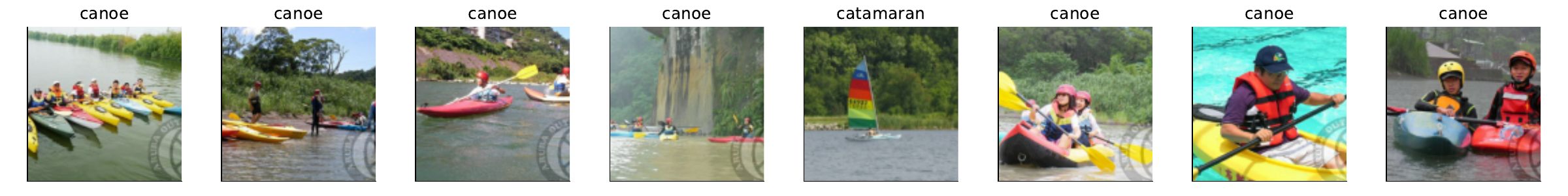}
    \includegraphics[width=\linewidth]{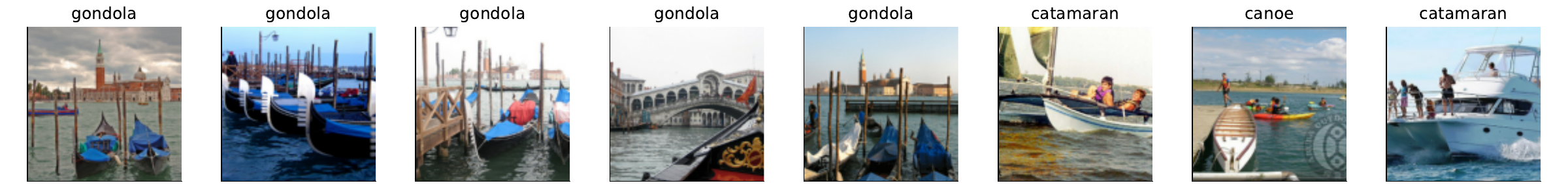}
    \vskip -1em
    \caption{Sample images from 4 clusters under `boat' uncovered in stage2\_block1 of a ResNet-50.}
\end{figure}

\begin{figure}
\centering  
    \includegraphics[width=\linewidth]{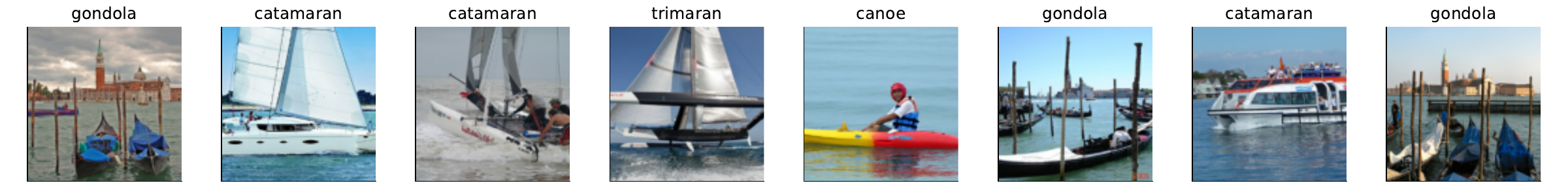}
    \includegraphics[width=\linewidth]{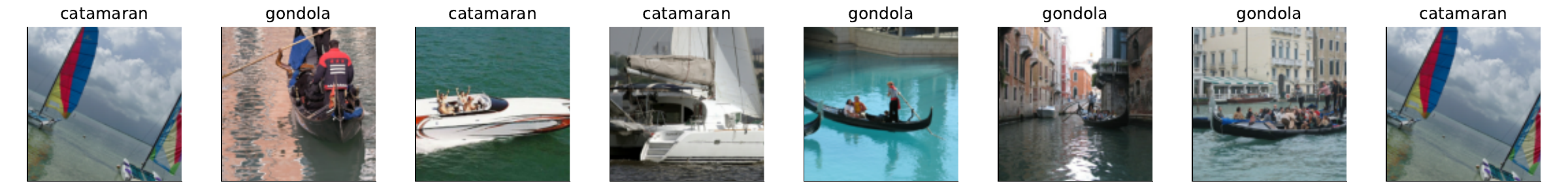}
    \includegraphics[width=\linewidth]{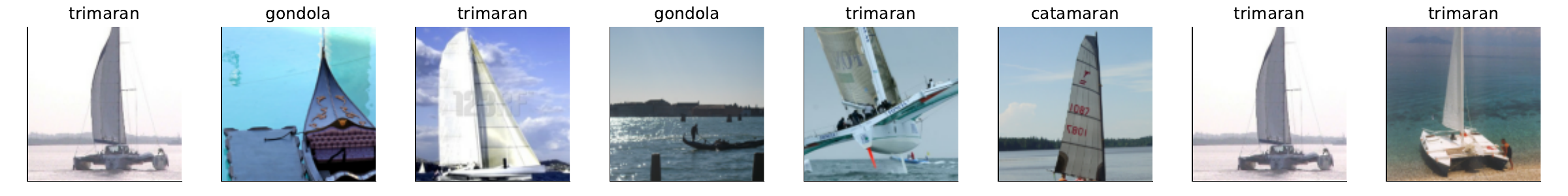}
    \includegraphics[width=\linewidth]{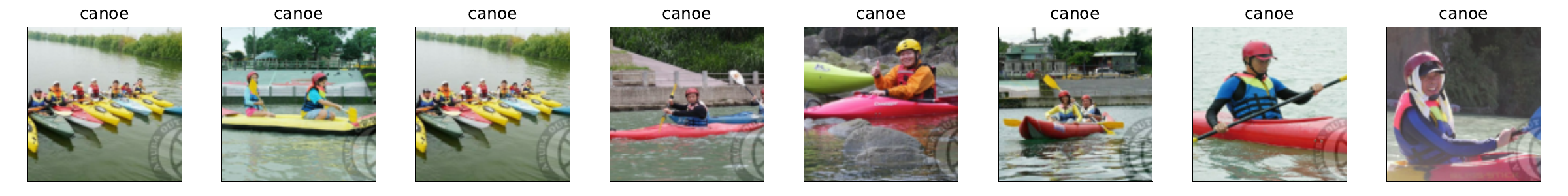}
    \vskip -1em
    \caption{Sample images from 4 clusters under `boat' uncovered in stage3\_block1 of a ResNet-50.}
\end{figure}

\begin{figure}
\centering  
    \includegraphics[width=\linewidth]{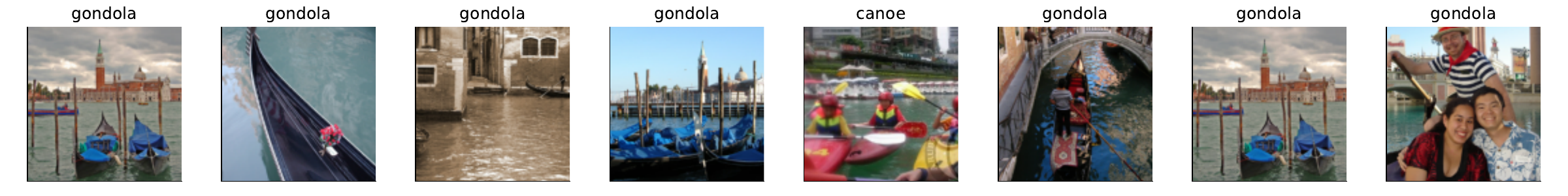}
    \includegraphics[width=\linewidth]{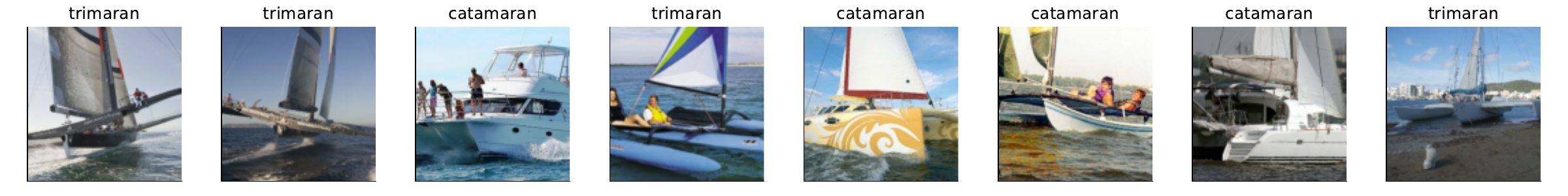}
    \includegraphics[width=\linewidth]{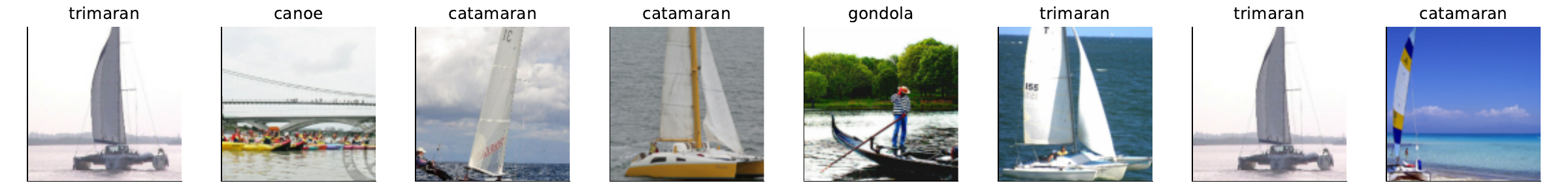}
    \includegraphics[width=\linewidth]{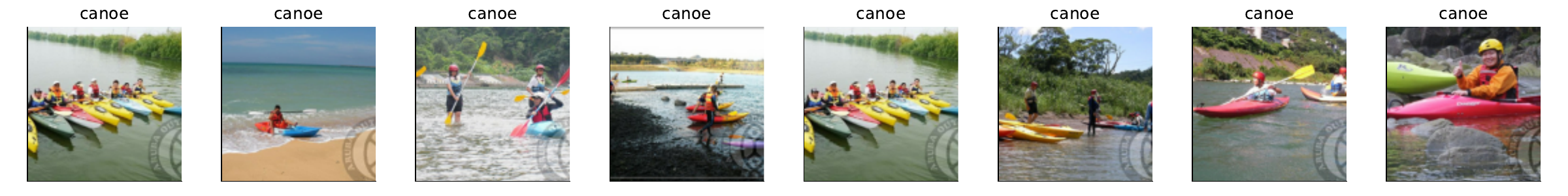}
    \vskip -0.5em
    \caption{Sample images from 4 clusters under `boat' uncovered in stage4\_block1 of a ResNet-50.}
\end{figure}

\begin{figure}[h]
\centering  
\includegraphics[width=\linewidth]{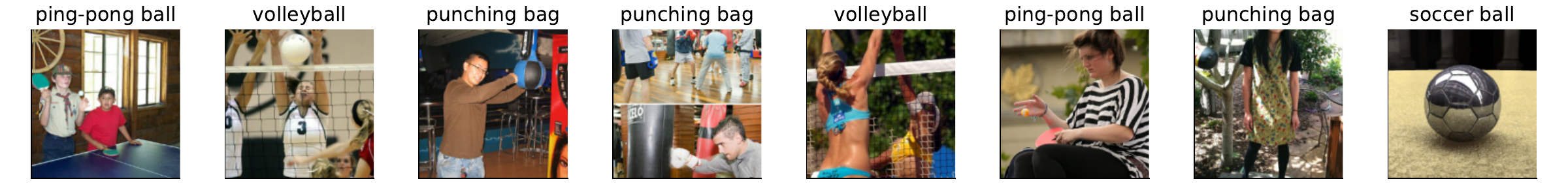}
\includegraphics[width=\linewidth]{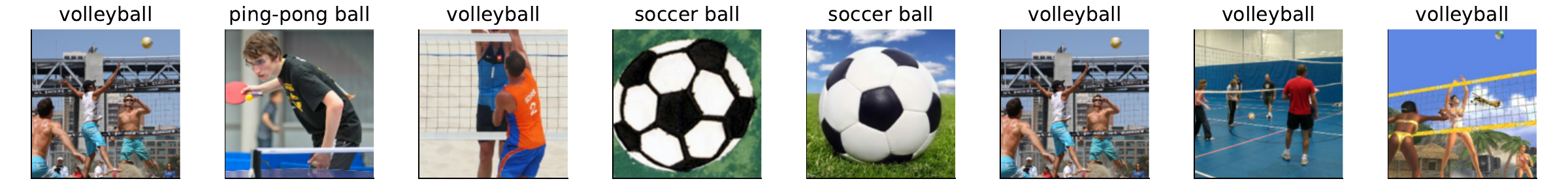}
\includegraphics[width=\linewidth]{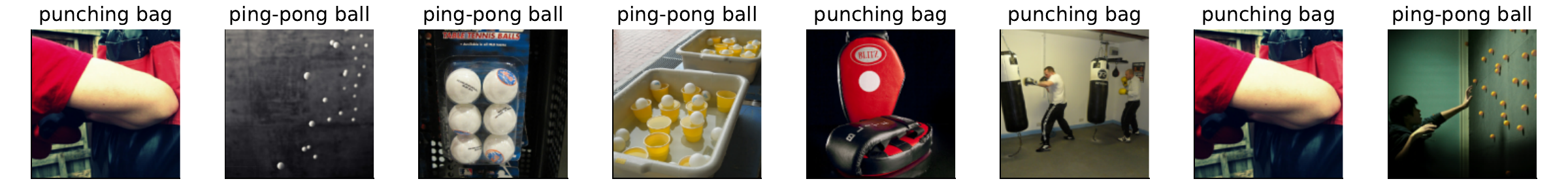}
\includegraphics[width=\linewidth]{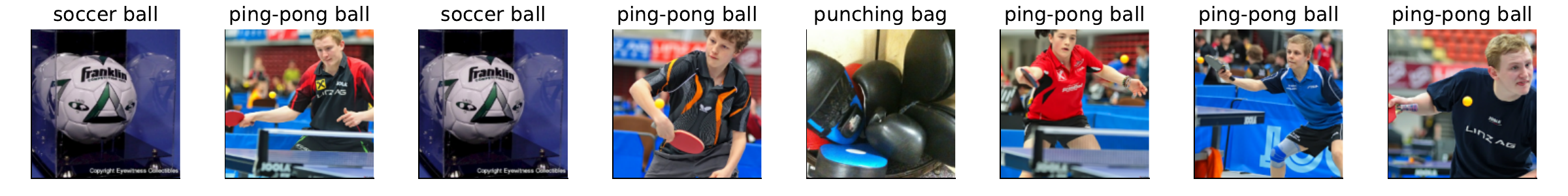}
\vskip -0.5em
\caption{Sample images from 4 clusters under `ball' uncovered in stage1\_block1 of a ResNet-50.}
\end{figure}

\begin{figure}
\centering
    \includegraphics[width=\linewidth]{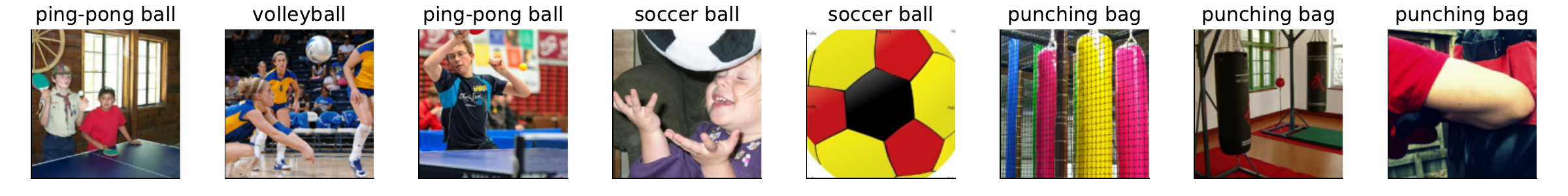}
    \includegraphics[width=\linewidth]{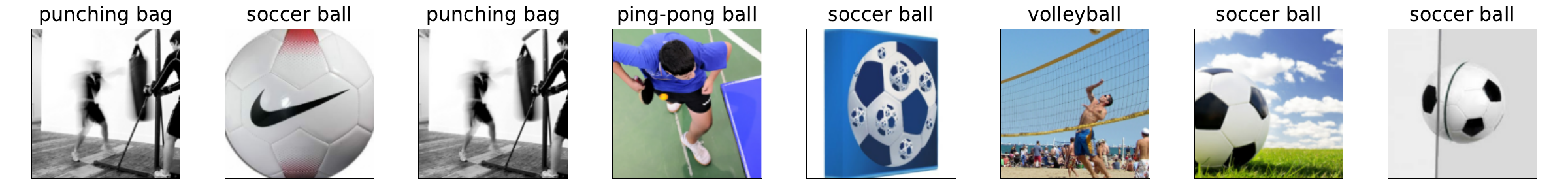}
    \includegraphics[width=\linewidth]{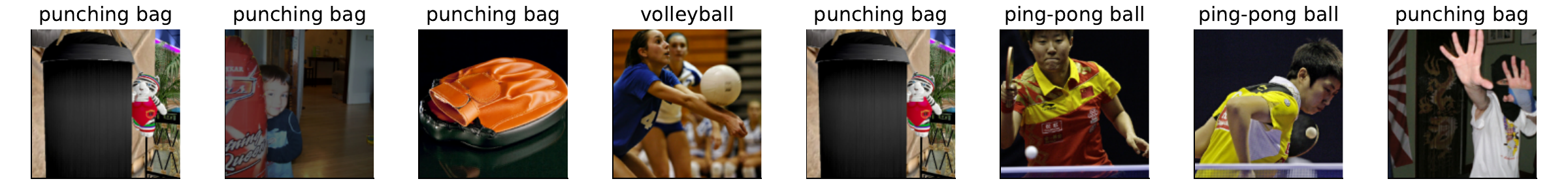}
    \includegraphics[width=\linewidth]{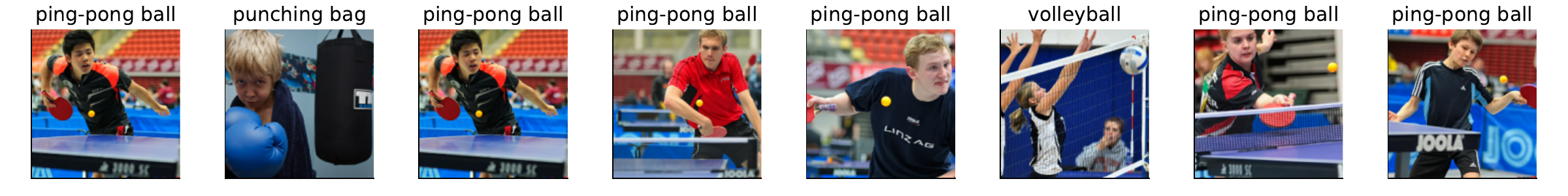}
    \vskip -1em
    \caption{Sample images from 4 clusters under `ball' uncovered in stage2\_block1 of a ResNet-50.}
\end{figure}

\begin{figure}
\centering  
    \includegraphics[width=\linewidth]{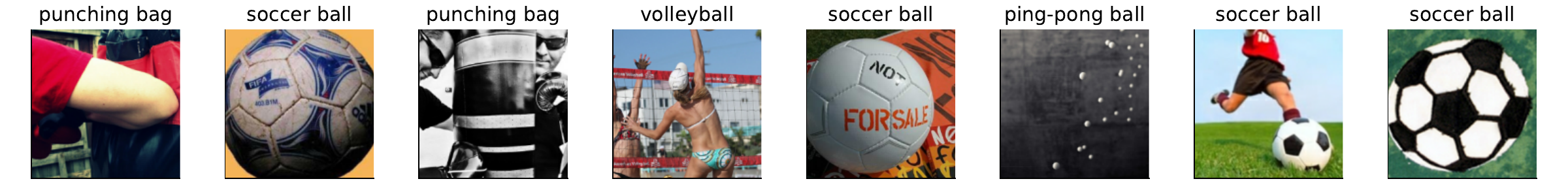}
    \includegraphics[width=\linewidth]{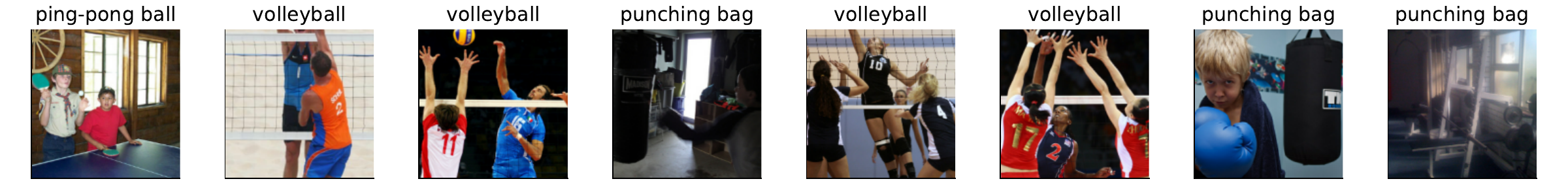}
    \includegraphics[width=\linewidth]{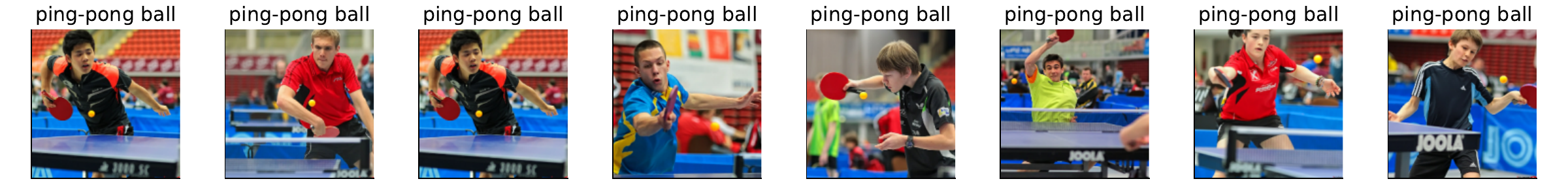}
    \includegraphics[width=\linewidth]{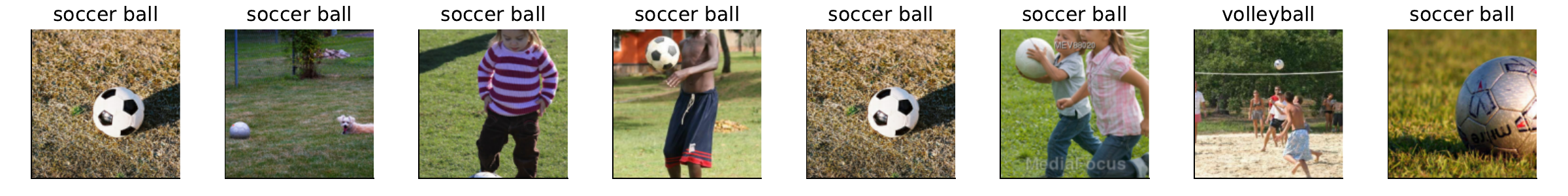}
    \vskip -1em
    \caption{Sample images from 4 clusters under `ball' uncovered in stage3\_block1 of a ResNet-50.}
\end{figure}

\begin{figure}
\centering  
    \includegraphics[width=\linewidth]{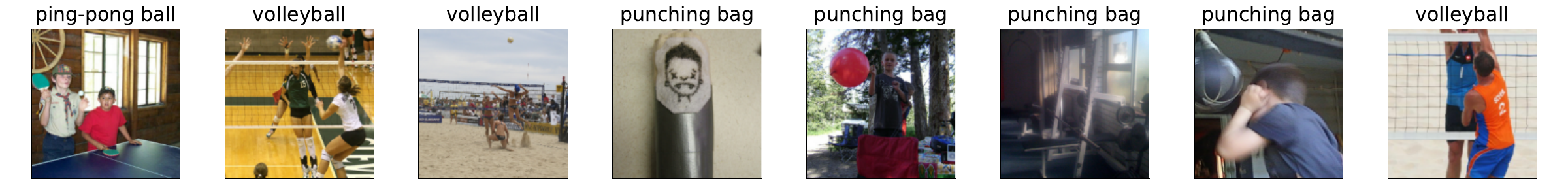}
    \includegraphics[width=\linewidth]{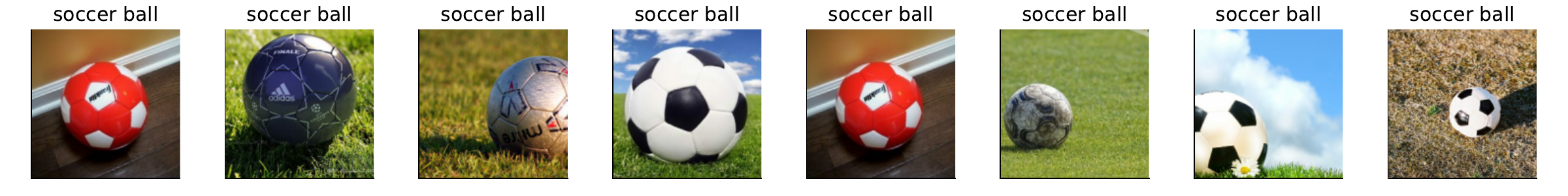}
    \includegraphics[width=\linewidth]{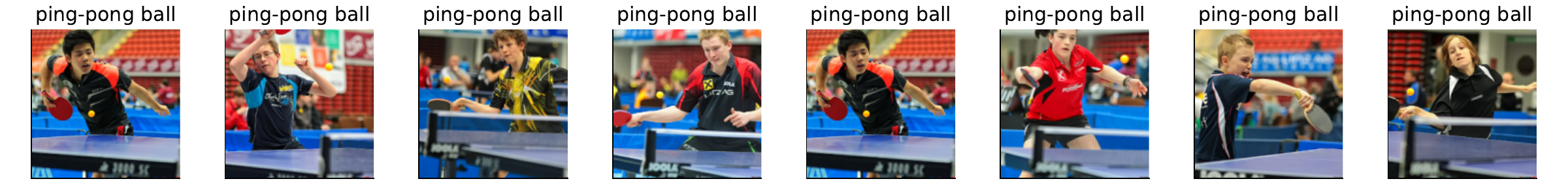}
    \includegraphics[width=\linewidth]{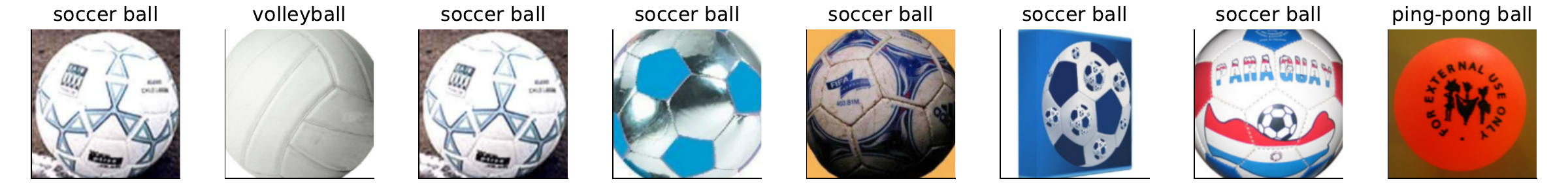}
    \vskip -1em
    \caption{Sample images from 4 clusters under `ball' uncovered in stage4\_block1 of a ResNet-50.}
\end{figure}

\begin{figure}[h]
\centering
\includegraphics[trim=0 0 0 0,clip,width=0.7\linewidth]
{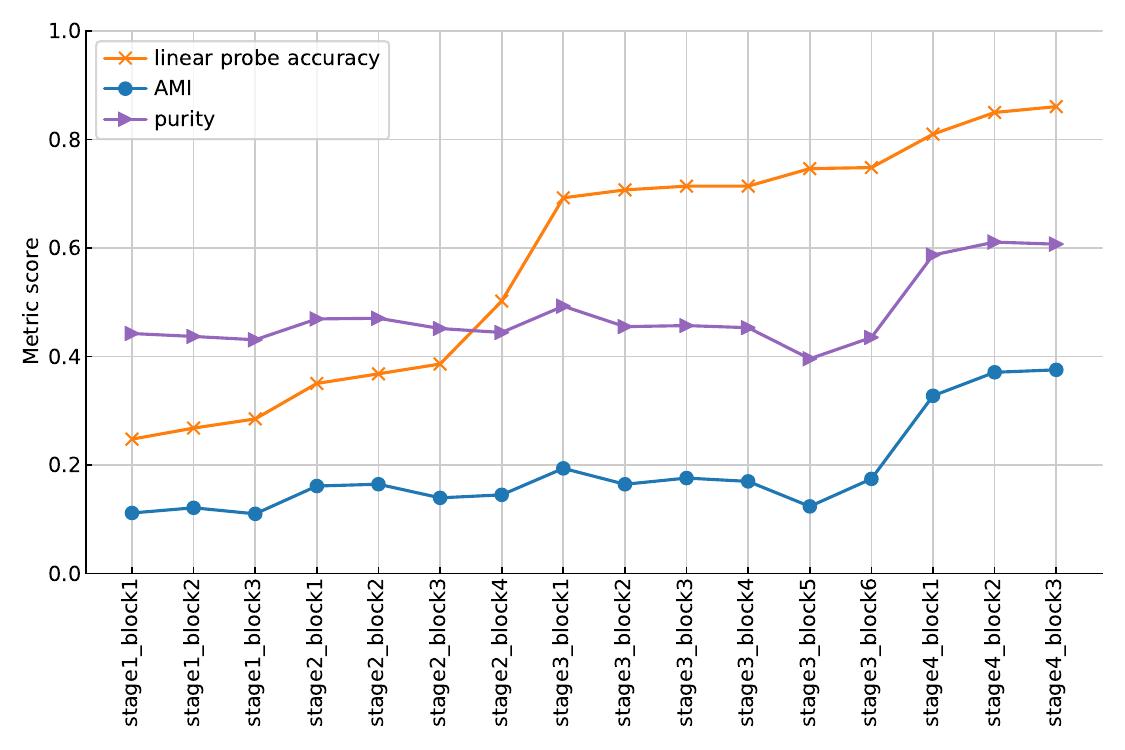}
\caption{We compare AMI and purity of subclass clustering to linear probe performance for different layers of a ResNet-50 trained on entity-13 dataset. Similar to what was observed in Figure \ref{fig:linear_probe} with entity-13-shuffled data, clustering metrics exhibit more variations across layers, and are not entirely aligned with linear probe accuracy.
}
\label{fig:linear_probe_more_results}
\end{figure}

\begin{figure}[h]
\subfloat{
    \includegraphics[trim=0 0 0 0,clip,width=0.49\linewidth]{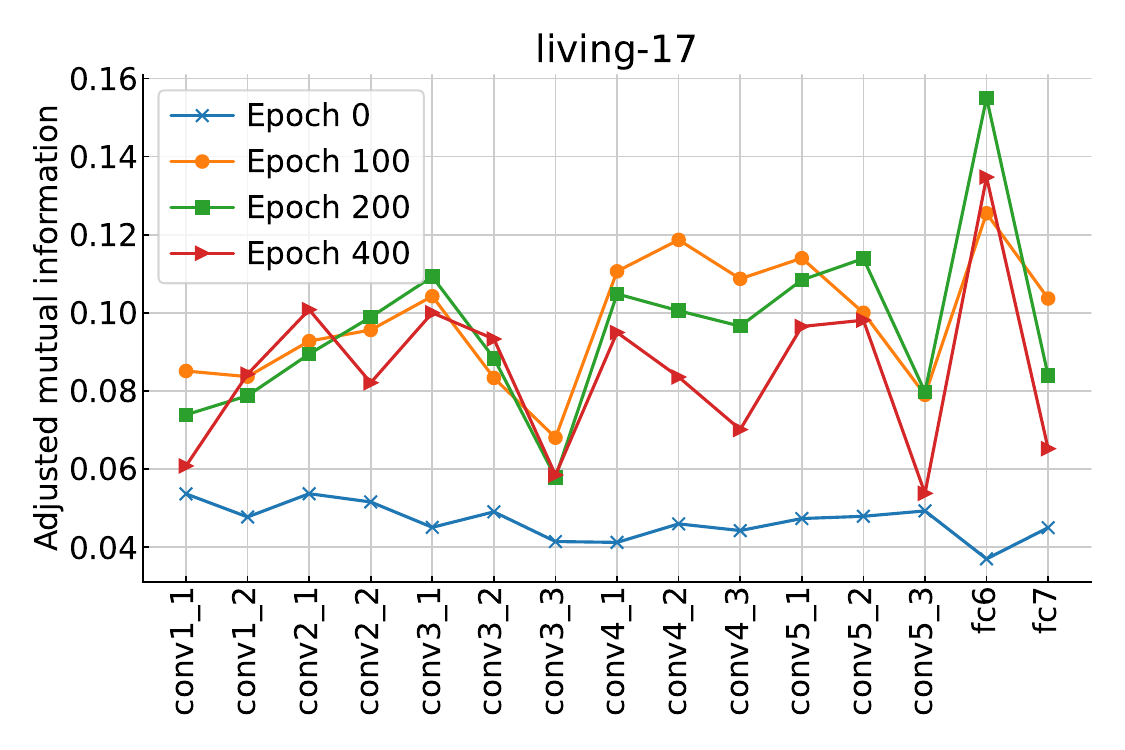}
}\hfill
\subfloat{
  \includegraphics[trim=0 0 0 0,clip,width=0.49\linewidth]{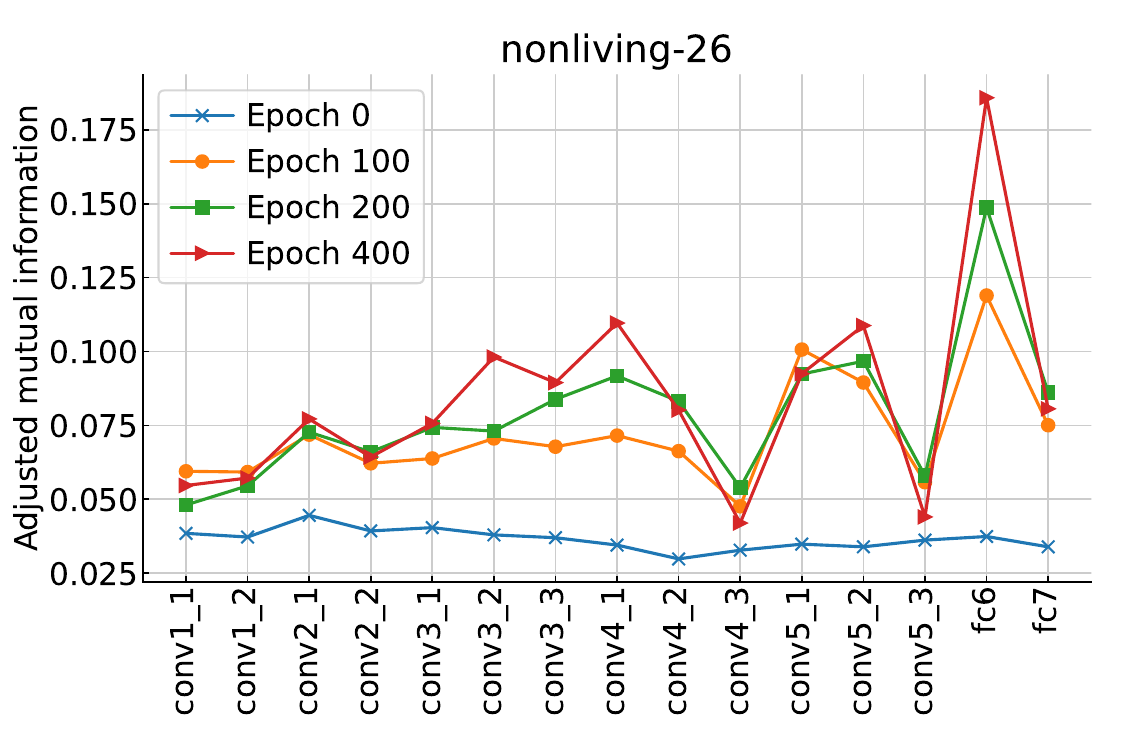}
}
\\\vspace{-1em}
\subfloat{
  \includegraphics[trim=0 0 0 0,clip,width=0.49\linewidth]{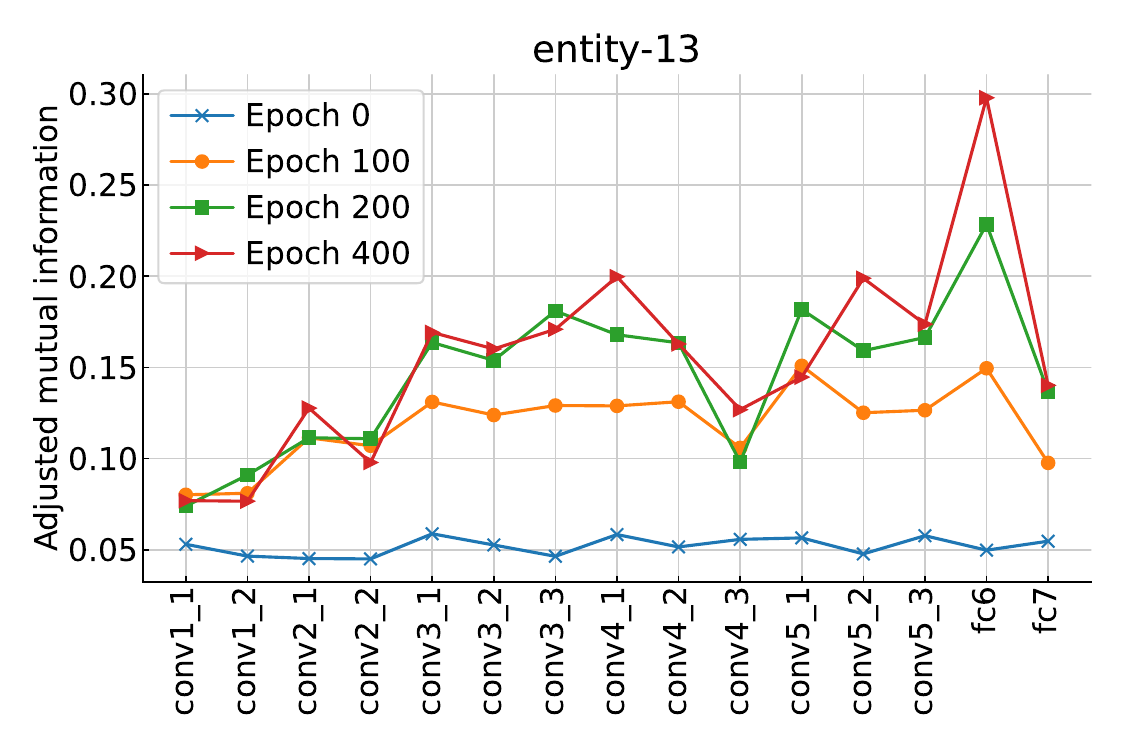}
}\hfill
\subfloat{
  \includegraphics[trim=0 0 0 0,clip,width=0.49\linewidth]{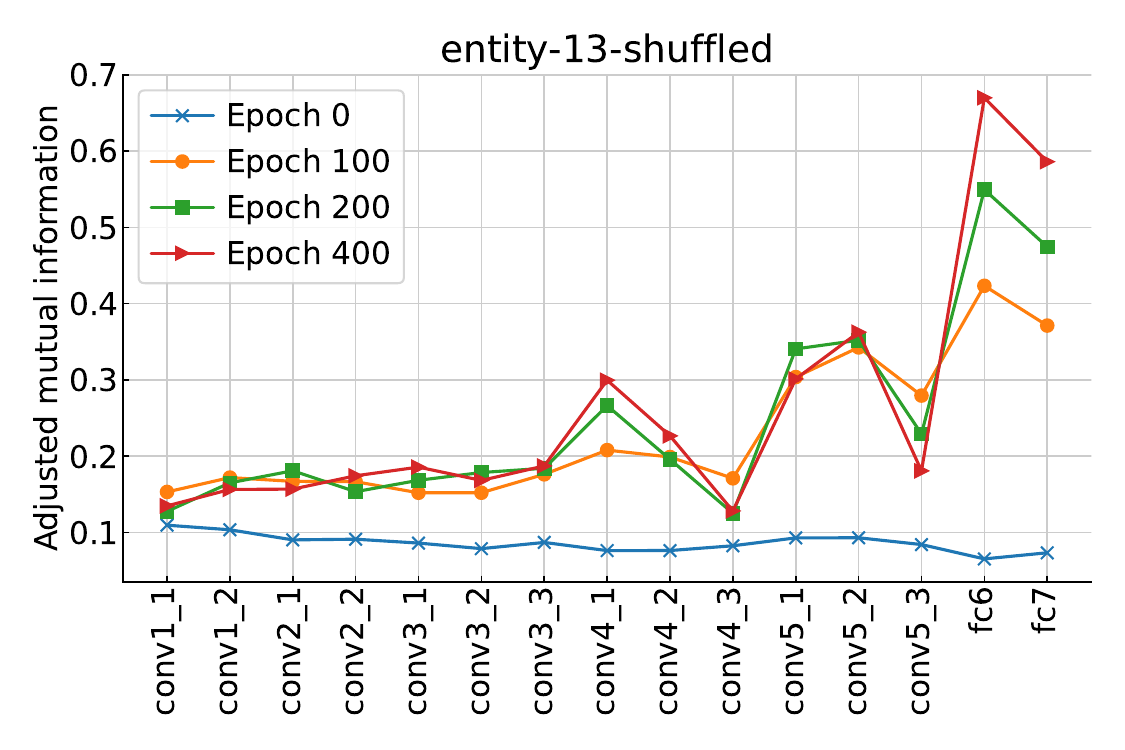}
}
\vskip 1em
\caption{We track AMI changes over time, across layers of VGG-16 models trained on different BREEDS datasets. We find that AMI values for most layers vary continuously throughout the course of training. The only exception is training with entity-13-shuffled dataset, where the AMI stays mostly unchanged for the majority of the network after epoch 100 (i.e., 25\% of total training duration), and only the clustering performance of the last few fully-connected layers continues to increase over time.}
\label{fig:ami_over_step_vgg}
\end{figure}

\FloatBarrier
\section{The role of training data} \label{app:data}
\begin{figure}[h]
\centering
\includegraphics[trim=0 0 0 0,clip,width=0.48\linewidth]
{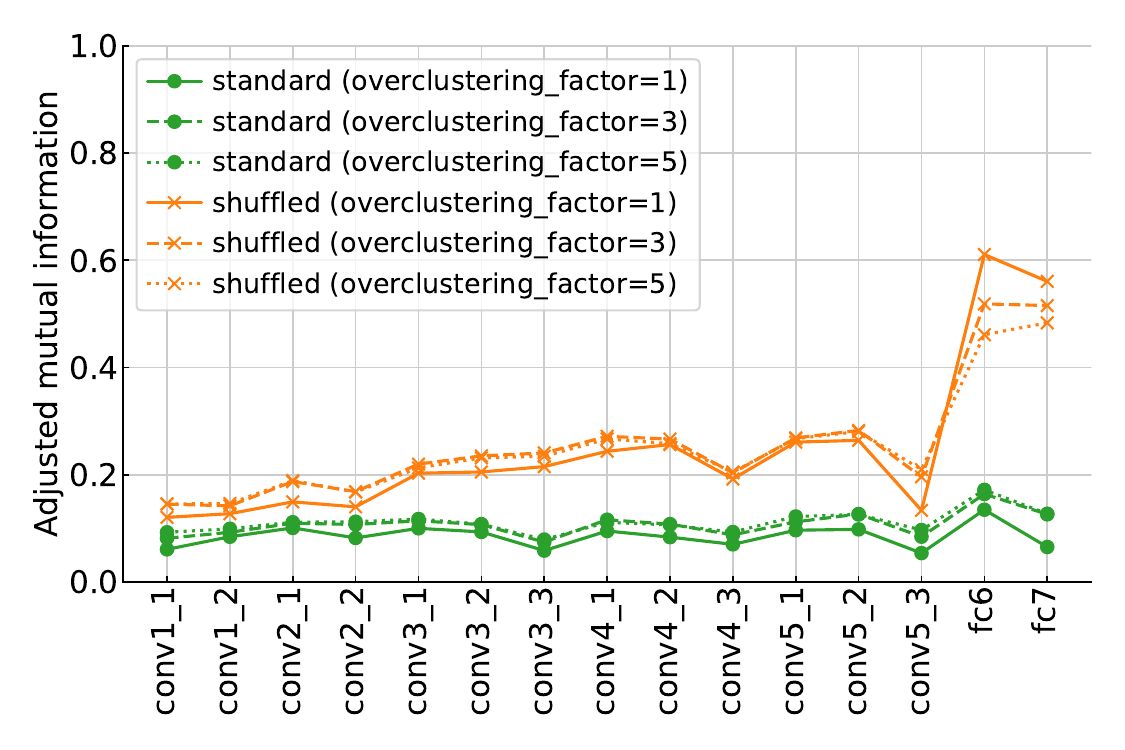}
\includegraphics[trim=0 0 0 0,clip,width=0.49\linewidth]
{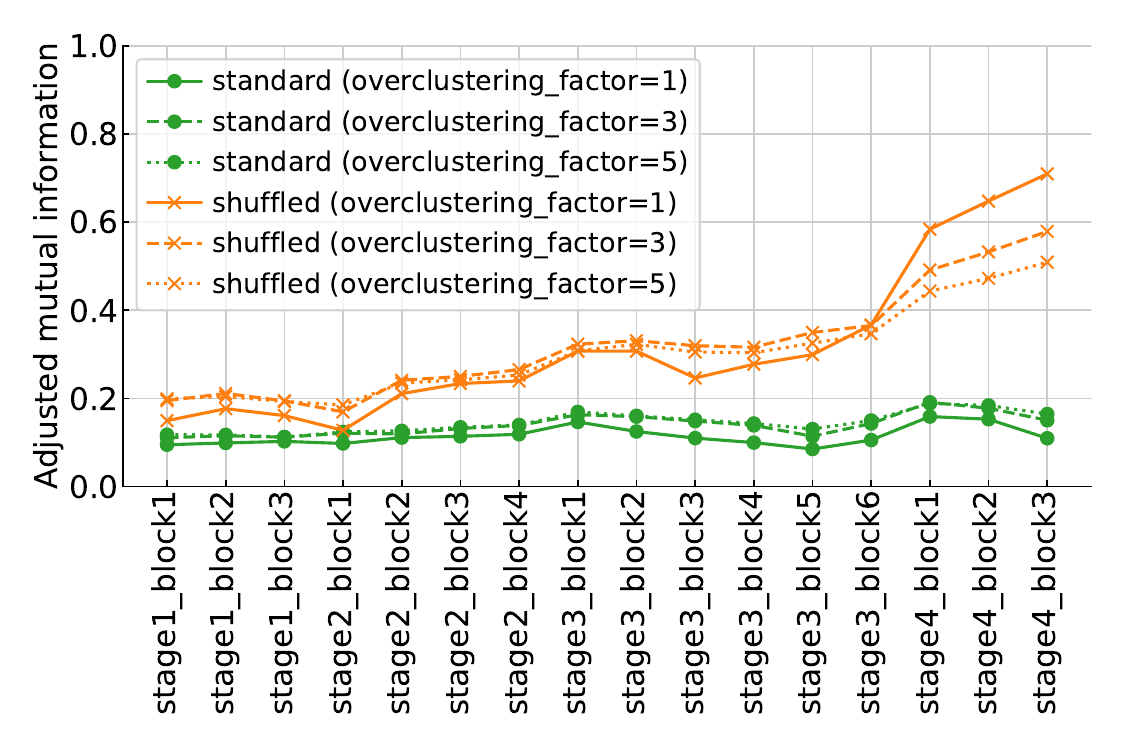}
\caption{We repeat experiment involving randomizing subclass-superclass mapping described in Section \ref{sec:shuffling} on living-17 dataset. Similar to what was observed in Figure \ref{fig:shuffle_vs_standard}, this intervention increases AMI across layers in both VGG-16 (left) and ResNet-50 (right). The improvements in clustering quality tend to be the largest near the end of either architecture.
}
\label{fig:shuffle_vs_standard_living17}
\end{figure}

\FloatBarrier
\vspace{2em}
\section{The role of loss function} 
\label{app:loss_functions}
\begin{figure}[h]
\includegraphics[trim=0 0 0 0,clip,width=\linewidth]
{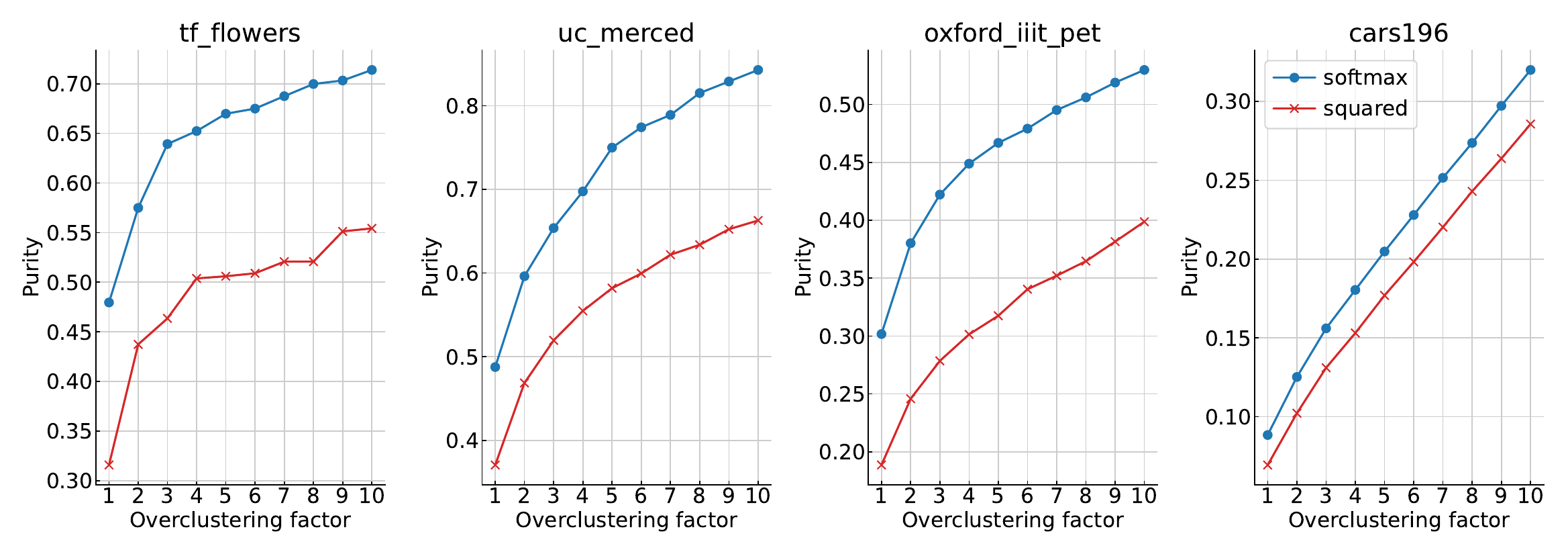}
\caption{On living-17 dataset, training a ResNet-50 with squared loss performs worse than training with softmax cross-entropy at recovering the class structures in out-of-distribution data. This trend holds across all four external datasets that we look at.
}
\label{fig:ood_squared_vs_softmax_full}
\vskip -1em
\end{figure}

\begin{figure}[h]
 \begin{minipage}{0.55\linewidth}
\centering
\includegraphics[trim=0 0 0 0,clip,width=\linewidth]
{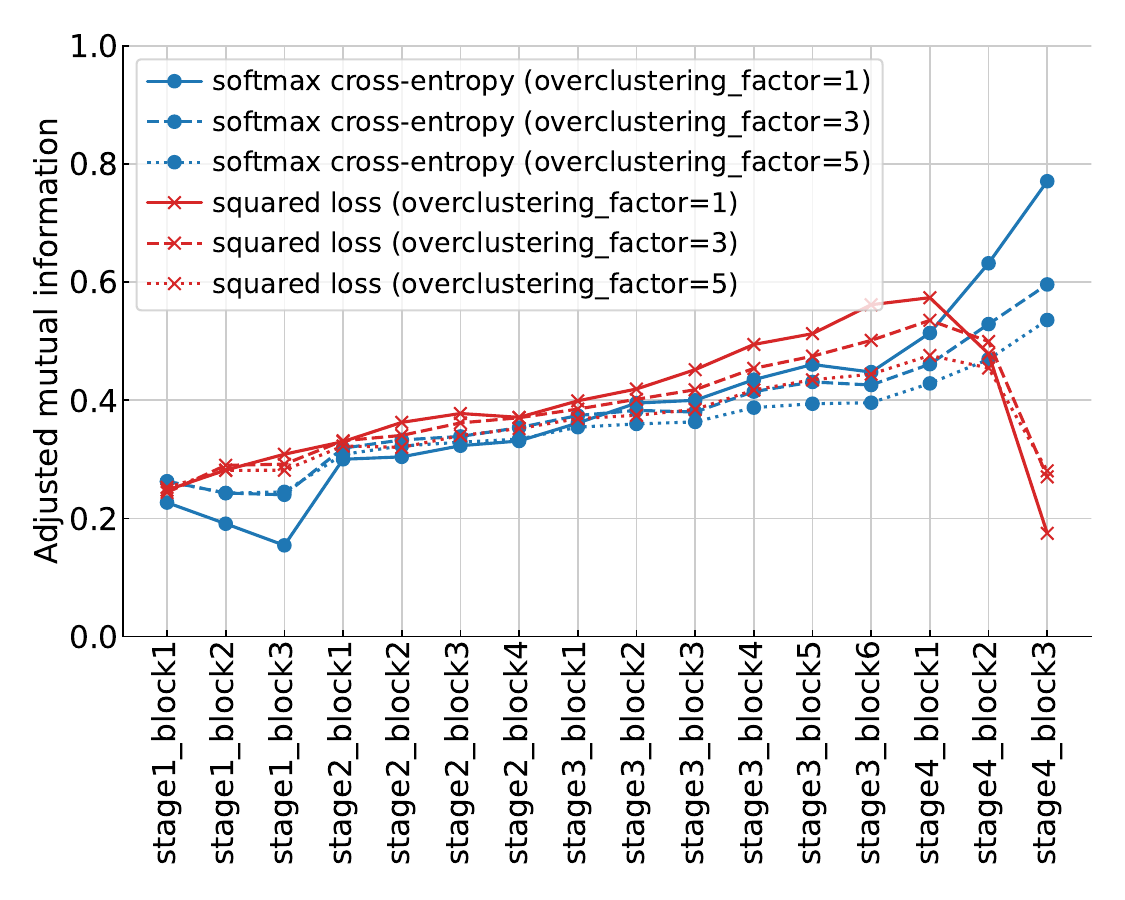}
\end{minipage}
 \begin{minipage}{0.43\linewidth}
\caption{Comparing AMI across layers for training a ResNet-50 with squared error vs softmax cross-entropy loss on entity-13-shuffled dataset. The two objective functions show similar AMI profiles for most of the network, except at the later layers where squared error sees a steep drop in AMI, possibly due to overfitting to the training classes \cite{kornblith2021better}.
\label{fig:squared_vs_softmax_entity13}
}
\end{minipage}
\vskip -1em
\end{figure}

\begin{figure}[h]
\includegraphics[trim=0 0 0 0,clip,width=\linewidth]
{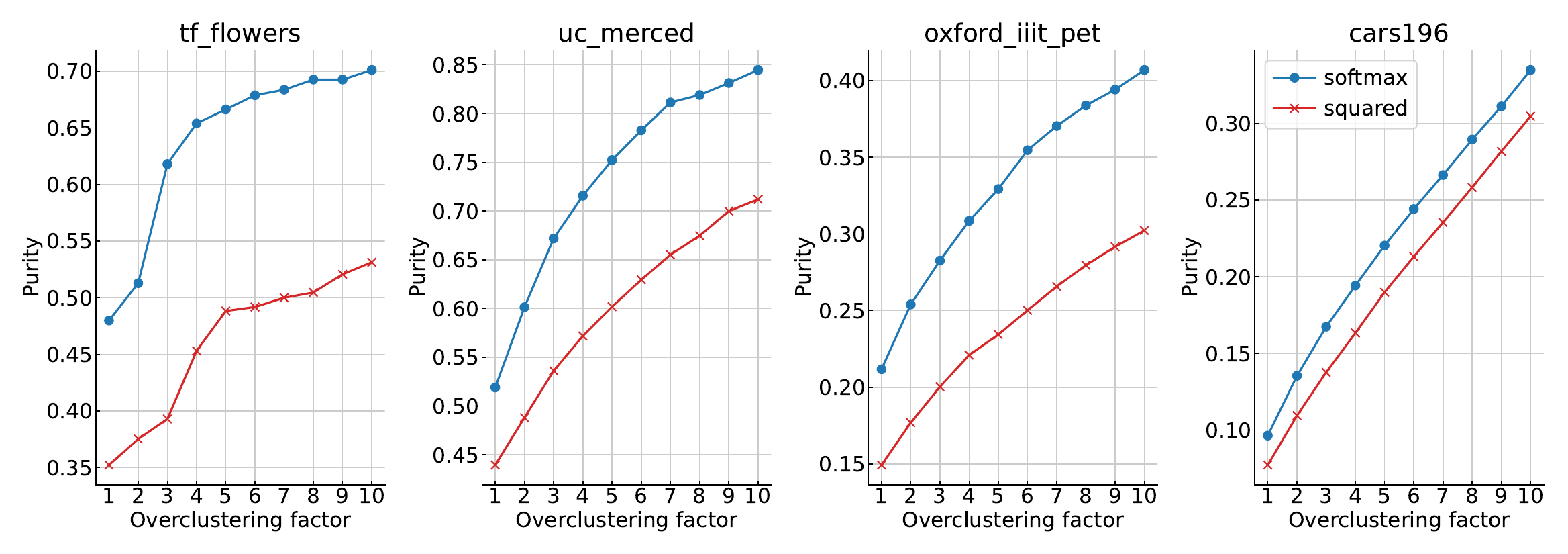}
\caption{On entity-13-shuffled dataset, training a ResNet-50 with squared loss performs worse than training with softmax cross-entropy at recovering the class structure in out-of-distribution data. This trend holds across all four external datasets that we look at.
}
\label{fig:ood_squared_vs_softmax_full_entity13}
\end{figure}

Figure \ref{fig:ood_squared_vs_softmax_full} shows the full results of performing clustering using the representations of a ResNet-50 trained with squared loss, on four external datasets, for the experiment described in Section \ref{sec:squared_vs_softmax}. Recall that this model was initially trained on living-17 dataset. We repeat the analysis for a ResNet-50 trained on entity-13-shuffled, and observe similar effects when using squared loss compared to using softmax cross-entropy. Refer to Figures \ref{fig:squared_vs_softmax_entity13} and \ref{fig:ood_squared_vs_softmax_full_entity13} for more details. 

\FloatBarrier
\section{The role of model architecture} \label{app:model_architecture}
\begin{figure}[ht!]
\begin{minipage}{0.55\linewidth}
\centering
\includegraphics[trim=0 0 0 0,clip,width=\linewidth]
{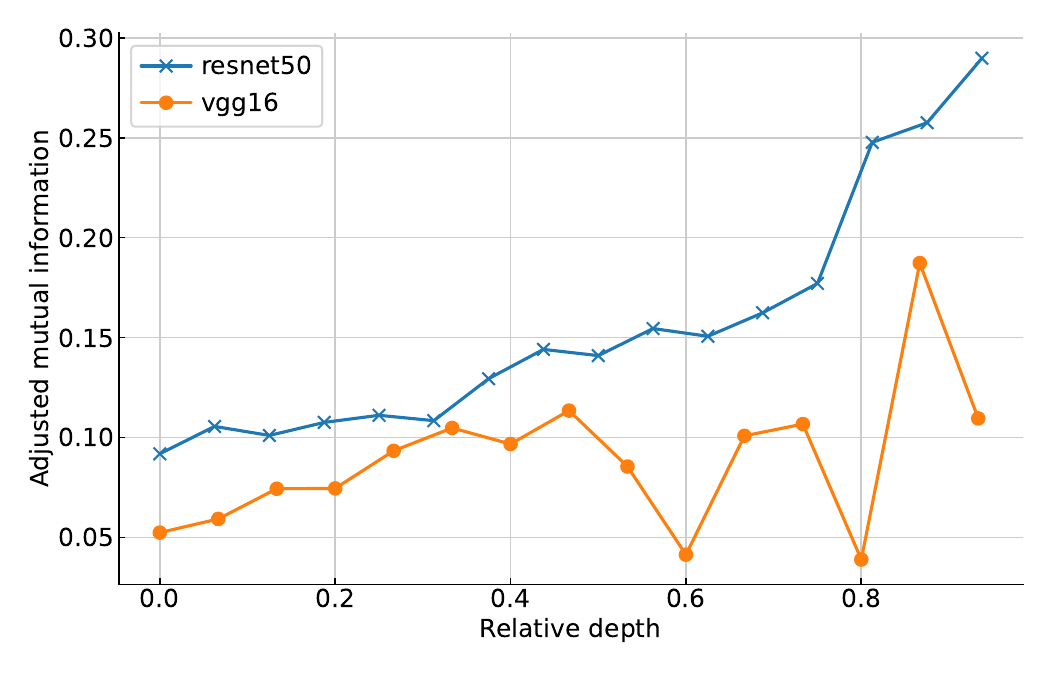}
\end{minipage}
\begin{minipage}{0.43\linewidth}
\caption{Similar to the observation in Figure \ref{sec:vgg_vs_resnet}, we find that the first fully connected layer (out of three) in VGG-16 is where maximum clustering performance is achieved. In contrast, clusterability tends to increase throughout and peak at the end of the network for ResNet-50 architecture.
}
\end{minipage}
\end{figure}

\begin{table}[h]
\begin{center}
\begin{tabular}{ | m{11em} | m{4.5em} | m{6em}| m{4em} | m{7.5em}| } 
  \hline
  Method & Living-17 & Nonliving-26 & Entity-13 & Entity-13-shuffled \\
  \hline
  ResNet-50 & 93.2 & 85.9 & 92.2 & 88.3 \\ 
  \hline
  VGG-16 & 83.1 & 53.4 & 86.3 & 76.5 \\ 
  \hline
  VGG-16 + Batch Norm & N/A & N/A & 86.6 & 78.1 \\
  \hline
  VGG-16 + Layer Norm & N/A & N/A & 77.2 & 
  73.6 \\
  \hline
  ViT-B/32 (average pooling) &  N/A & N/A & 81.3 & 72.6 \\
  \hline
\end{tabular}
\vskip 1em
\caption{Superclass classification accuracy of different model architectures trained with the BREEDS datasets used in our experiments. Reported results are averaged across 3 seeds.}
\label{tab:acc}
\end{center}
\vskip -1.5em
\end{table}


\FloatBarrier
\section{Comparing clusterings across networks} \label{app:across_seeds}
\vspace{-1.5em}
\begin{figure}[ht!]
\subfloat{
    \includegraphics[trim=0 0 0 0,clip,width=0.49\linewidth]{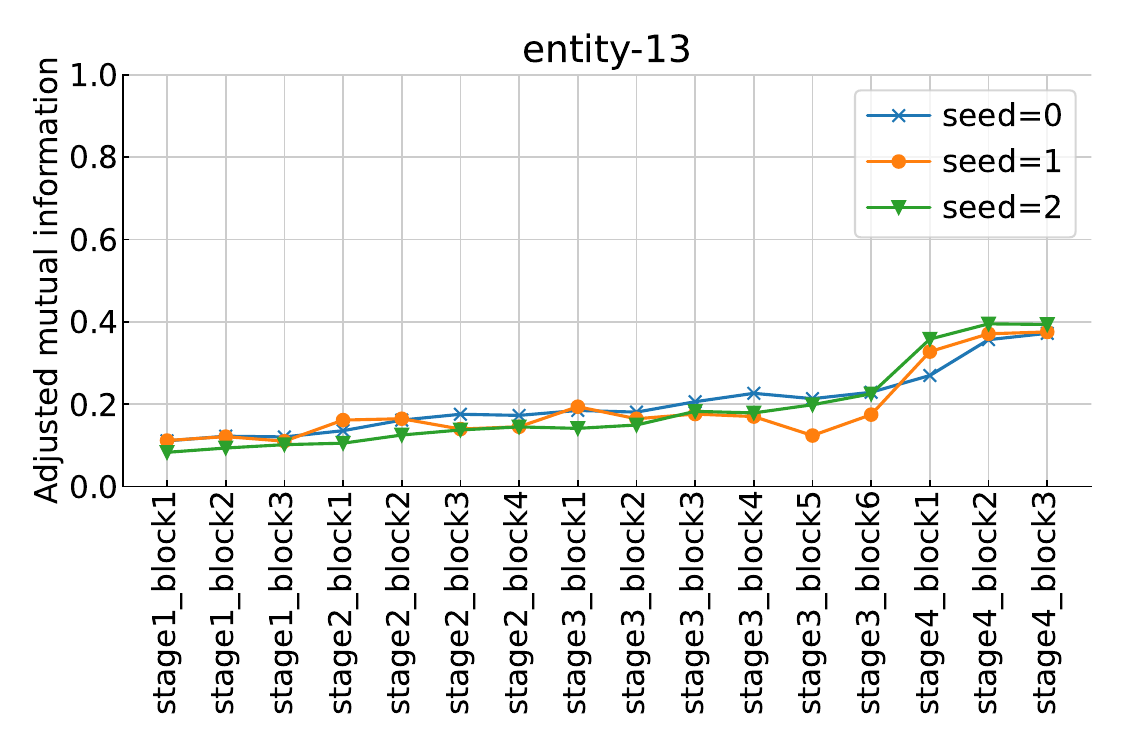}
}\hfill
\subfloat{
  \includegraphics[trim=0 0 0 0,clip,width=0.49\linewidth]{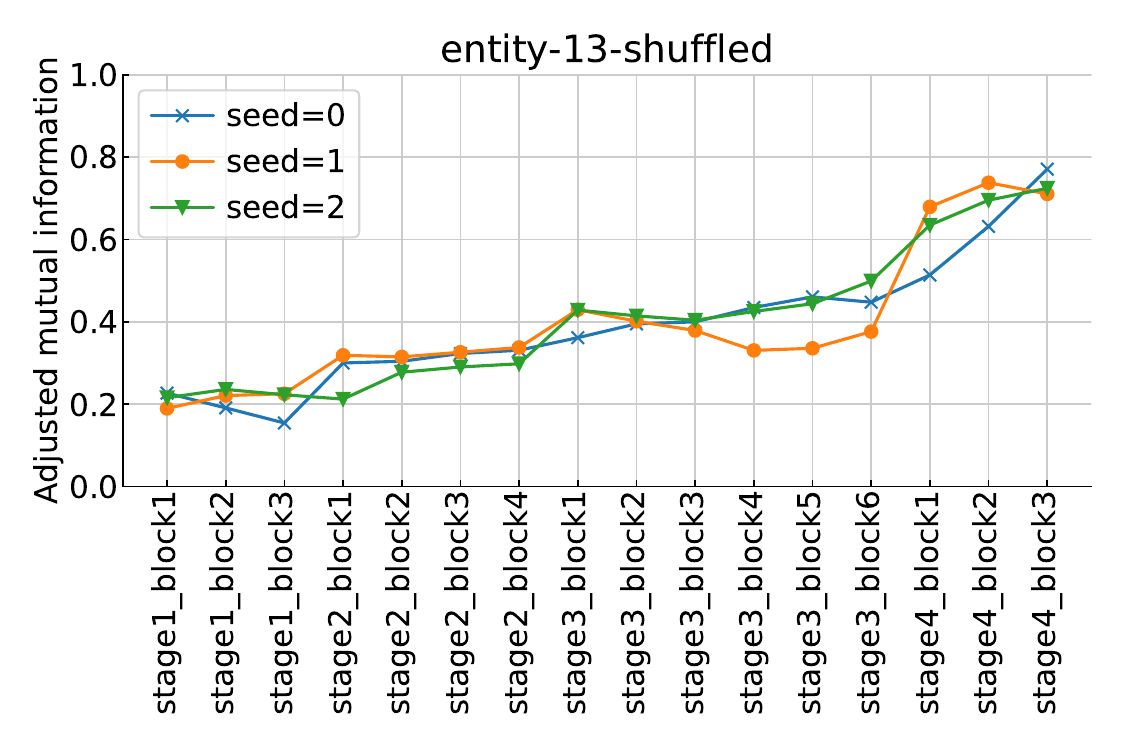}
}
\\ \vskip -1.5em
\subfloat{
  \includegraphics[trim=0 0 0 0,clip,width=0.49\linewidth]{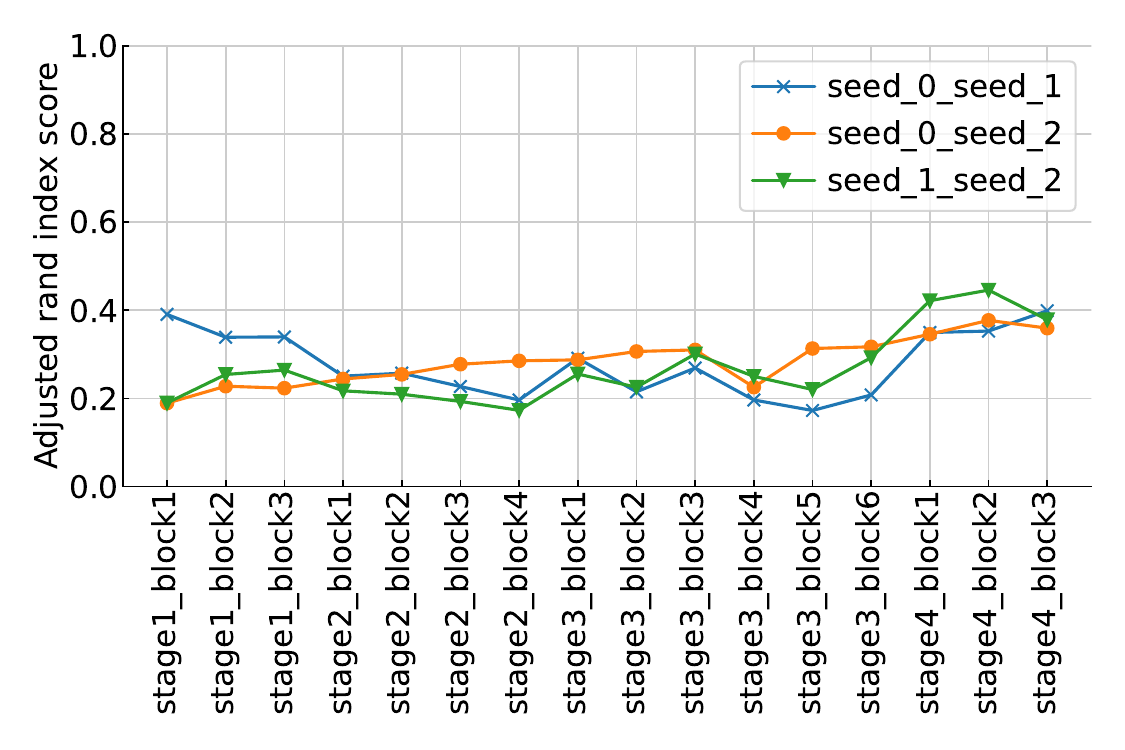}
}\hfill
\subfloat{
  \includegraphics[trim=0 0 0 0,clip,width=0.49\linewidth]{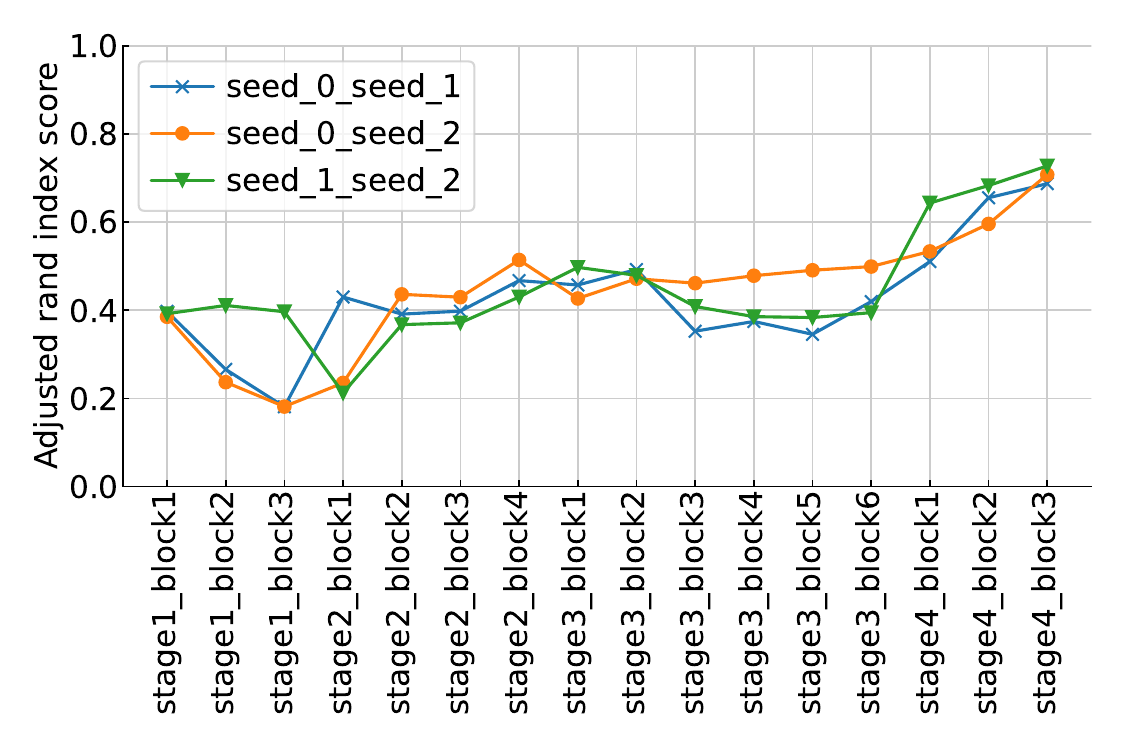}
}
\caption{Here we repeat the experiment described in Figure \ref{fig:ARI_across_seeds_vgg16} for ResNet-50 architecture. Top row shows the AMI across layers for three different training runs of ResNet-50 on entity-13 (left) and entity-13-shuffled (right). Similar to Figure \ref{fig:ARI_across_seeds_vgg16}, we observe that the random seeds produce similar AMI profiles. However, when measuring how aligned the actual clusters are across runs with adjusted Rand index, we find that the ARI values are relatively low for most layers (bottom row) --- the images assigned to each cluster vary substantially across runs. An exception is high cluster consistency found in the last few layers of a model trained on entity-13-shuffled, where we also observe a corresponding rise in AMI.
}
\label{fig:ARI_across_seeds_resnet50}
\vskip -1em
\end{figure}

\begin{figure}[]
\centering
\includegraphics[trim=0 0 0 0,clip,width=0.48\linewidth]
{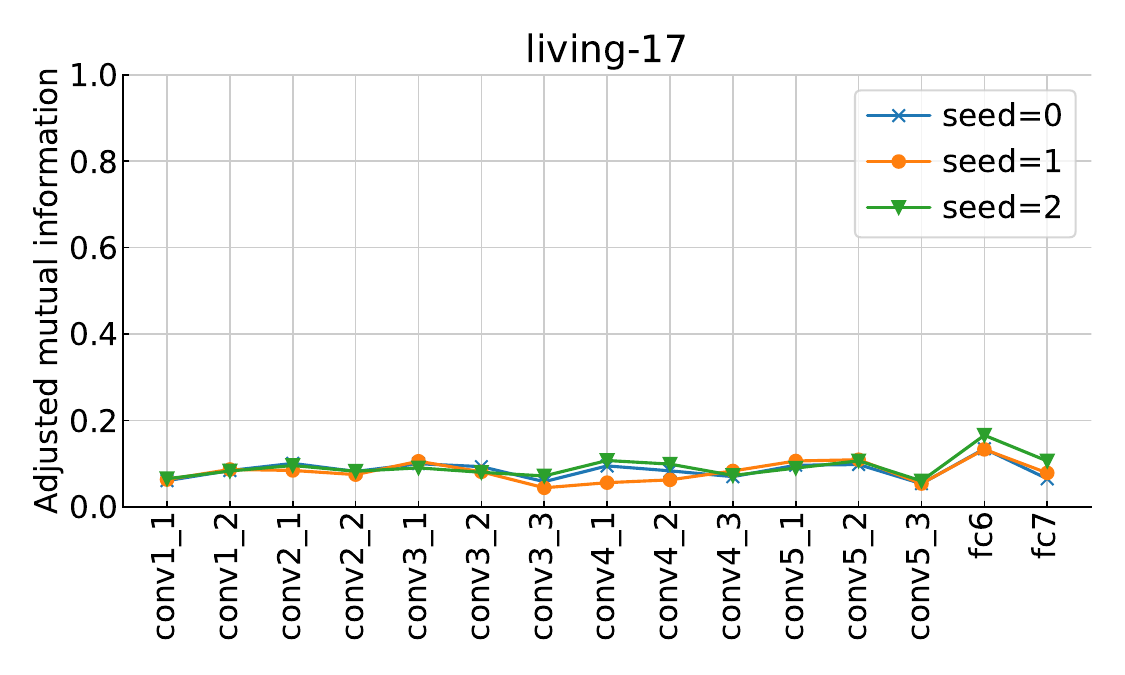}
\includegraphics[trim=0 0 0 0,clip,width=0.48\linewidth]
{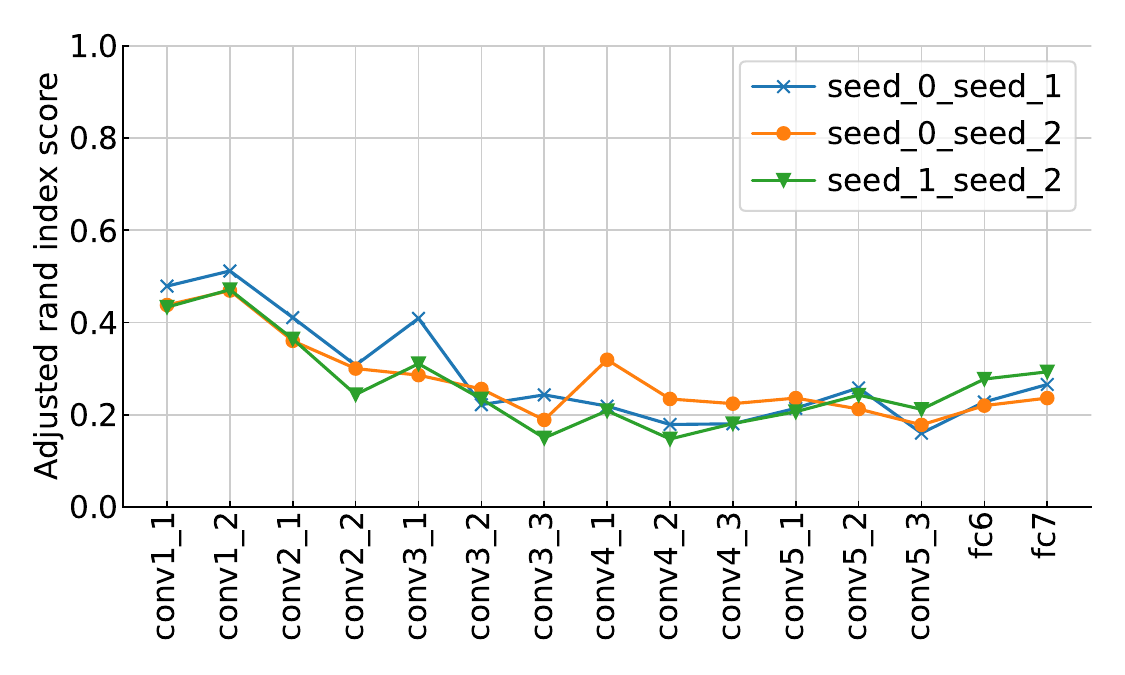}
\caption{We repeat the analysis described in Section \ref{sec:across_networks} for three different runs of a VGG-16 trained on living-17 dataset. Again, we find that the random seeds exhibit very similar AMI profiles, but the cluster assignments between them are highly inconsistent, with the peak consistency occurring in the early convolutional layers.
}
\label{fig:vgg_across_seeds}
\vskip -1em
\end{figure}

\begin{figure}[]
\centering
\includegraphics[trim=0 0 0 0,clip,width=0.47\linewidth]
{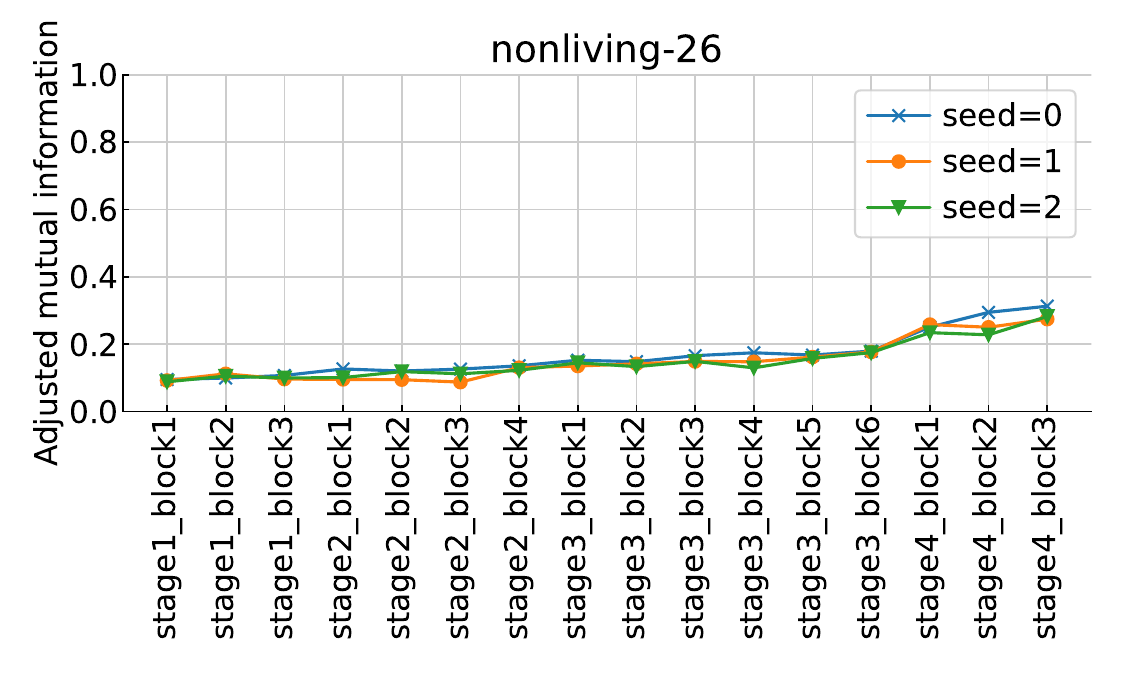}
\includegraphics[trim=0 0 0 0,clip,width=0.47\linewidth]
{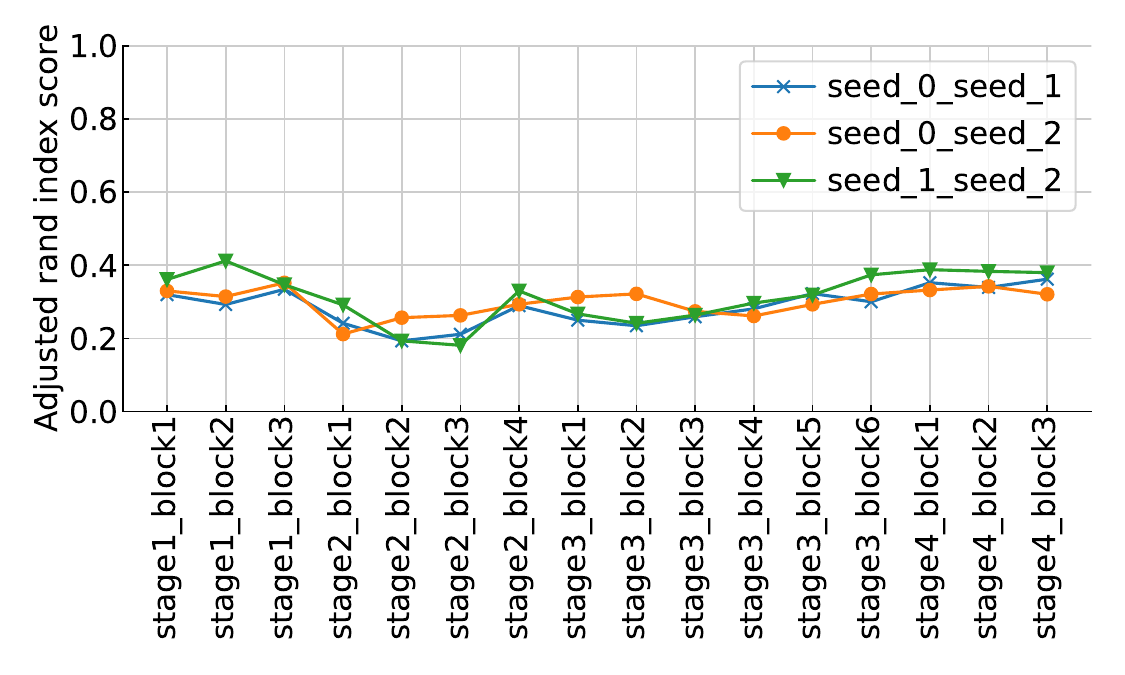}
\caption{We repeat the analysis described in Section \ref{sec:across_networks} for three different runs of a ResNet-50 trained on nonliving-26 dataset. Again, we find that the random seeds exhibit very similar AMI profiles, but the cluster assignments between them are highly inconsistent, with the consistency being comparatively high both at the start and at the end of the ResNet-50 network.
}
\label{}
\vskip -1em
\end{figure}

\begin{figure}[]
\centering
\includegraphics[trim=0 0 0 0,clip,width=0.48\linewidth]
{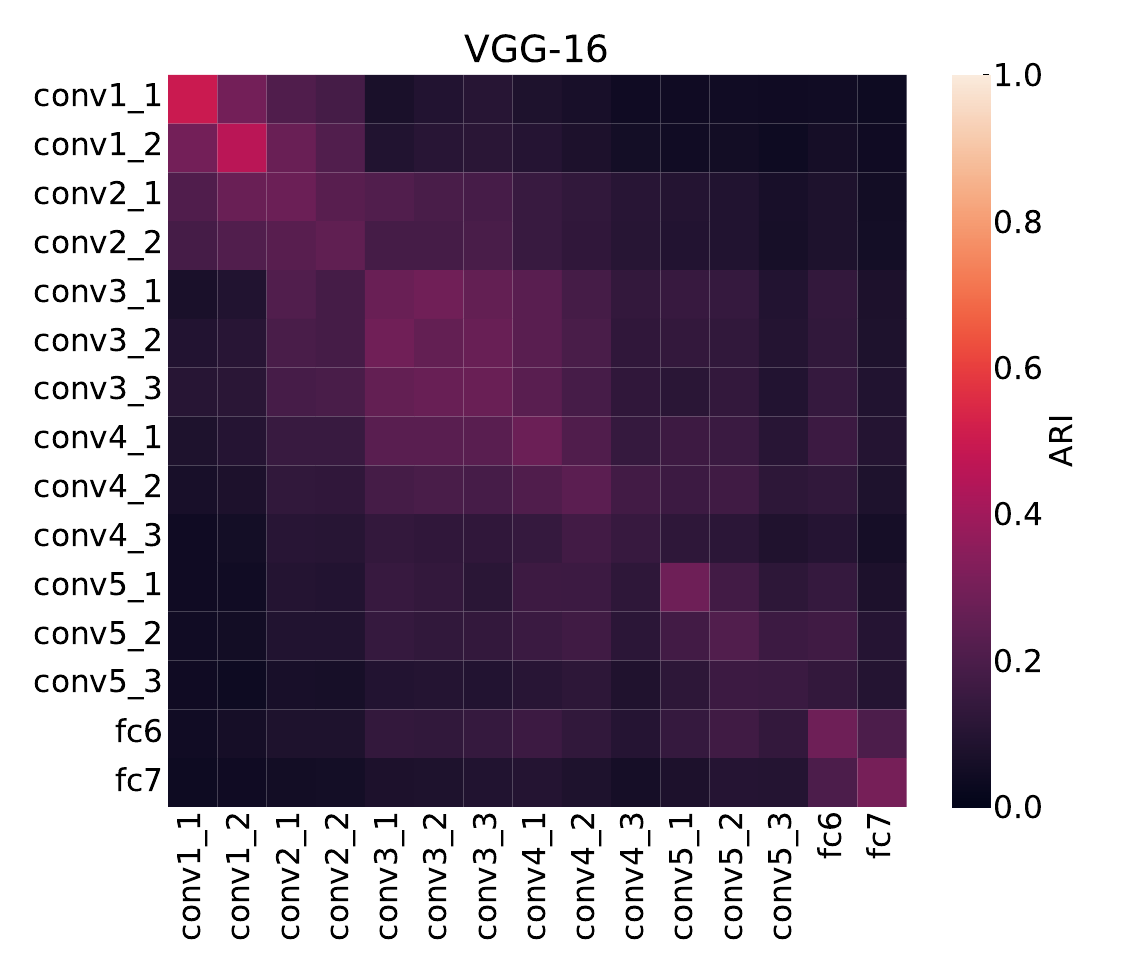}
\includegraphics[trim=0 0 0 0,clip,width=0.48\linewidth]
{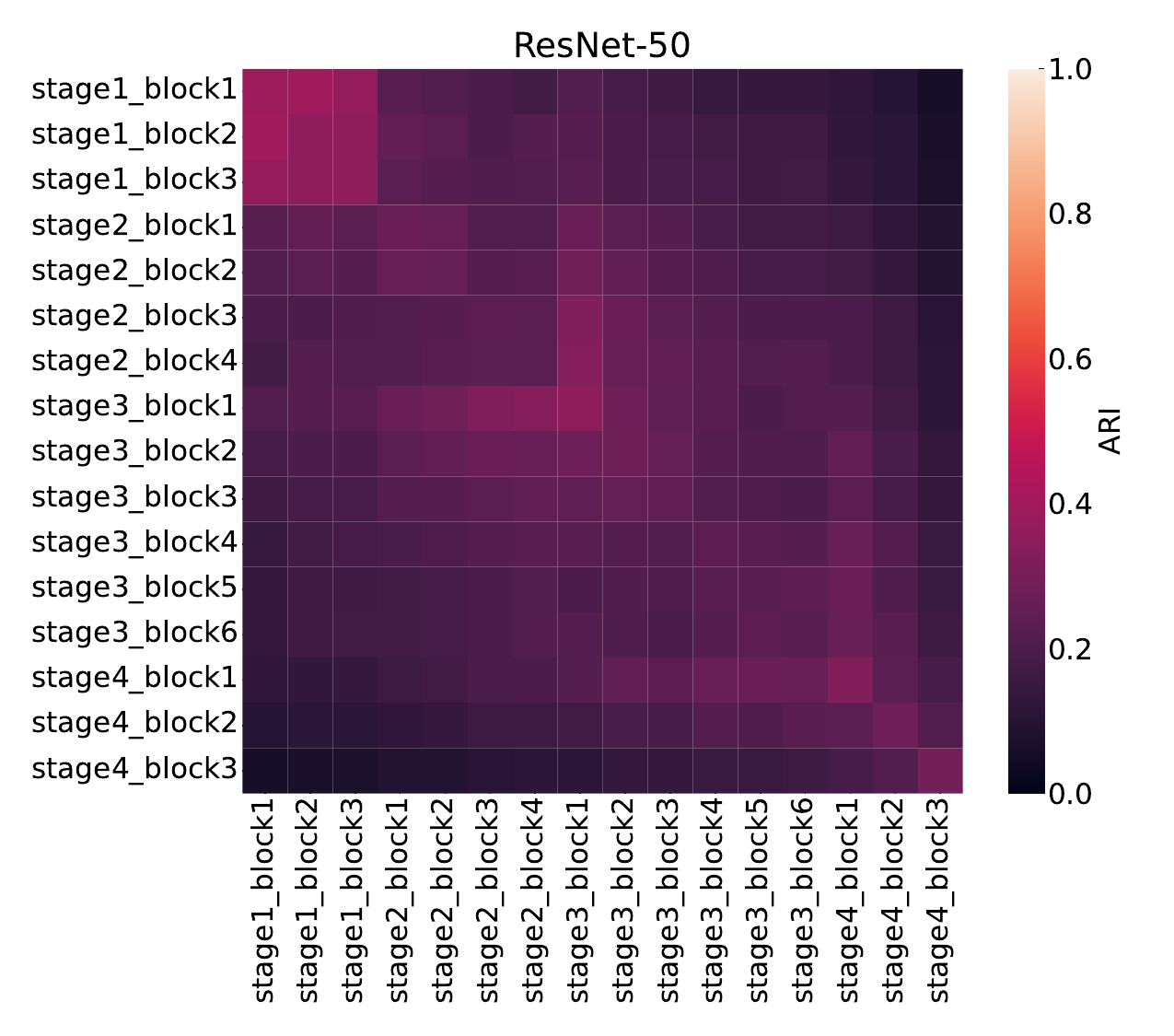}
\caption{We compare clustering consistency, measured with adjusted Rand index, on entity-13 for every pair of layers found across two training runs of the same model. We observe that in VGG-16 (left pane), corresponding layers between two random seeds show relatively higher consistency compared to other pairs of layers, though the ARI values are still low overall. However, this is not the case for ResNet-50 (right pane), where pairs of layers from the same `stage' can exhibit higher cluster alignment compared to corresponding layers.
}
\label{fig:entity13_4_subclasses_heatmap}
\vskip -1em
\end{figure}

\newpage
Below we show sample images from each of the 4 clusters uncovered under the superclass `ball' of nonliving-26, using embeddings from layer stage4\_block3 of ResNet-50s trained with different random initializations on the same dataset. Each row represents one cluster, and each image visualization is labeled with the ground-truth subclass name:
\begin{figure}[h]
\centering  
\includegraphics[width=\linewidth]{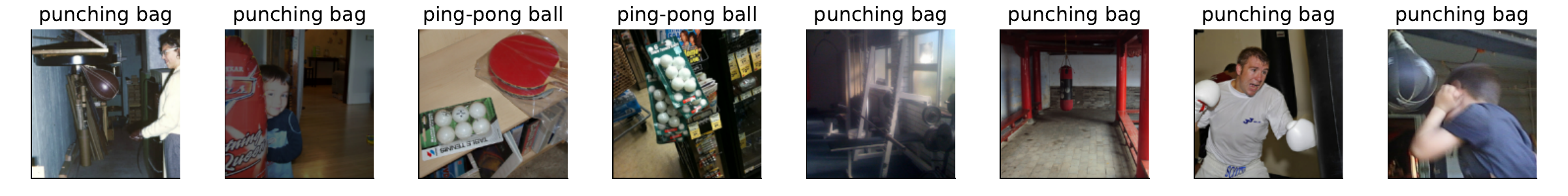}
\includegraphics[width=\linewidth]{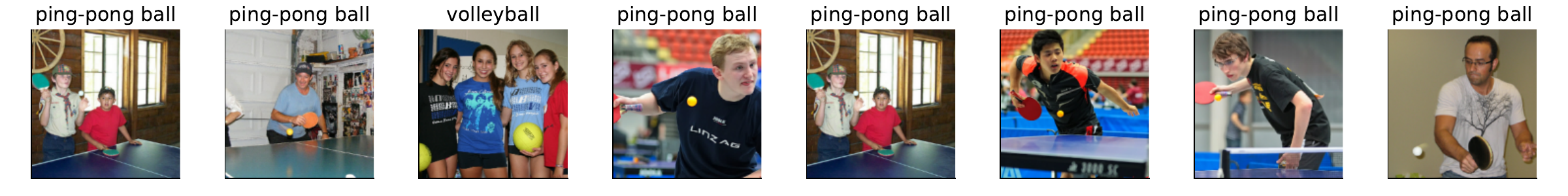}
\includegraphics[width=\linewidth]{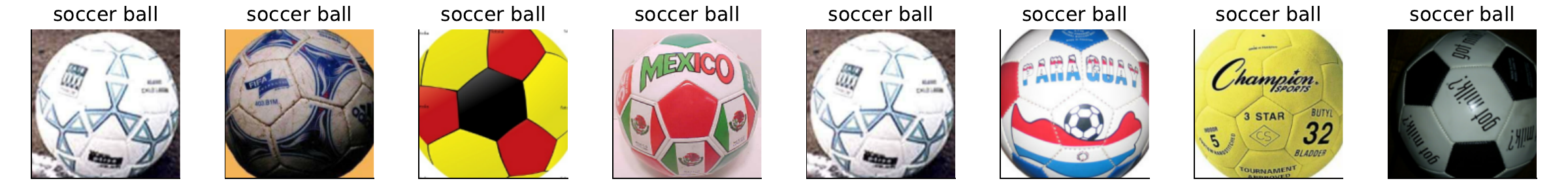}
\includegraphics[width=\linewidth]{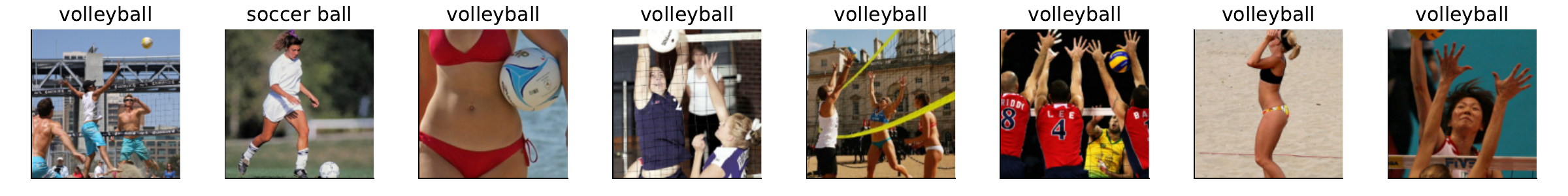}
\vskip -1em
\caption{Sample images from 4 clusters under `ball' uncovered in stage4\_block3 of ResNet-50 random seed 1.}
\end{figure}

\begin{figure}
\centering
\includegraphics[width=\linewidth]{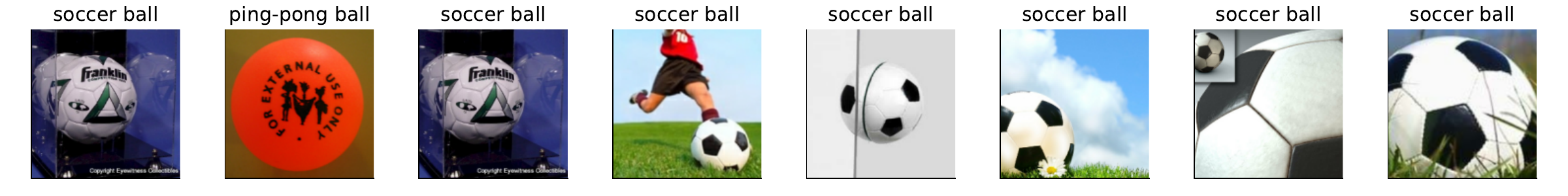}
\includegraphics[width=\linewidth]{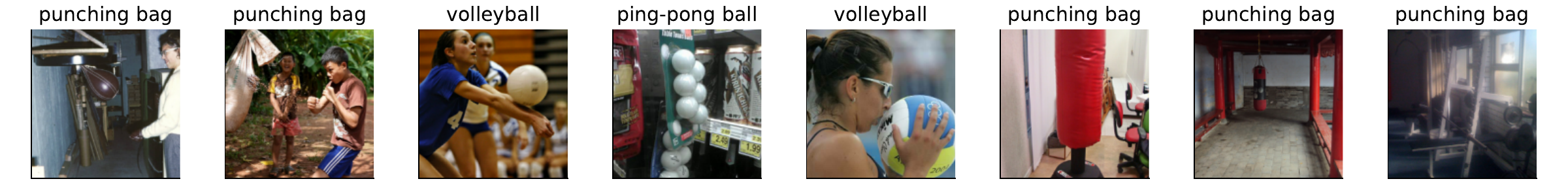}
\includegraphics[width=\linewidth]{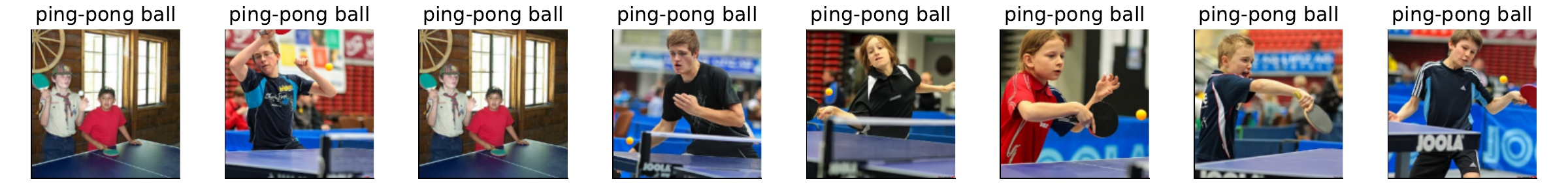}
\includegraphics[width=\linewidth]{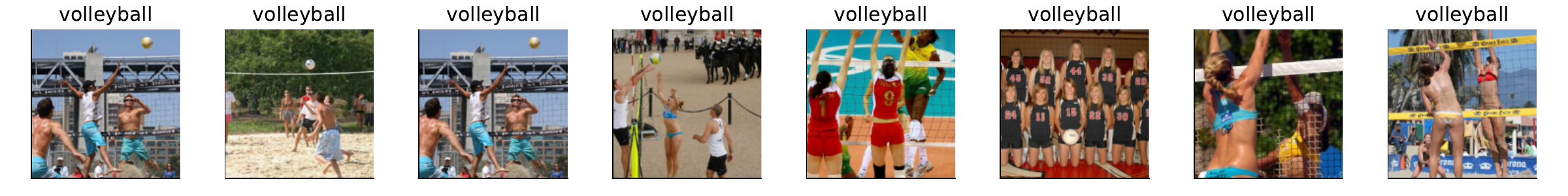}
\vskip -1em
\caption{Sample images from 4 clusters under `ball' uncovered in stage4\_block3 of ResNet-50 random seed 2.}
\end{figure}

\begin{figure}
\centering  
\includegraphics[width=\linewidth]{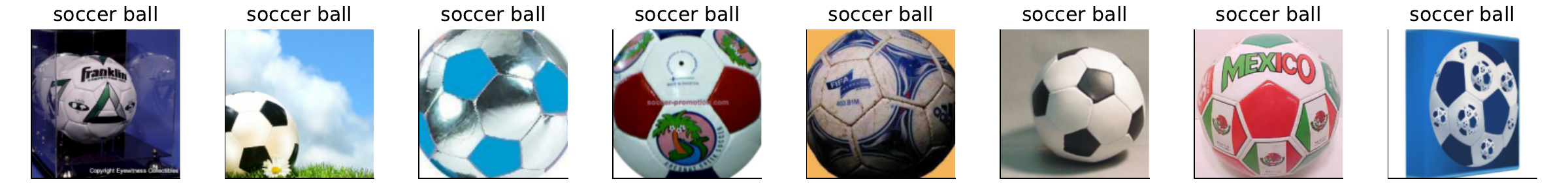}
\includegraphics[width=\linewidth]{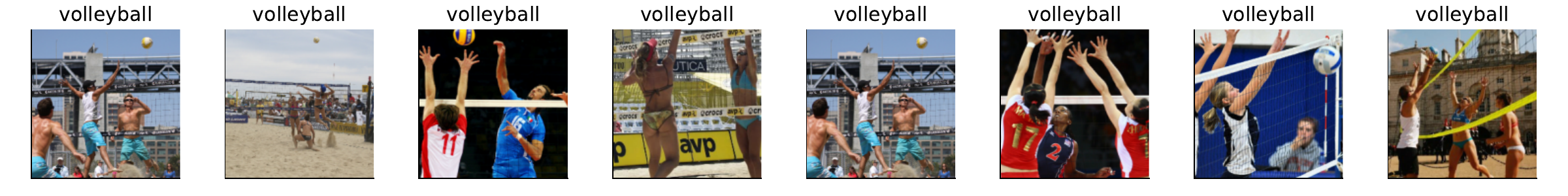}
\includegraphics[width=\linewidth]{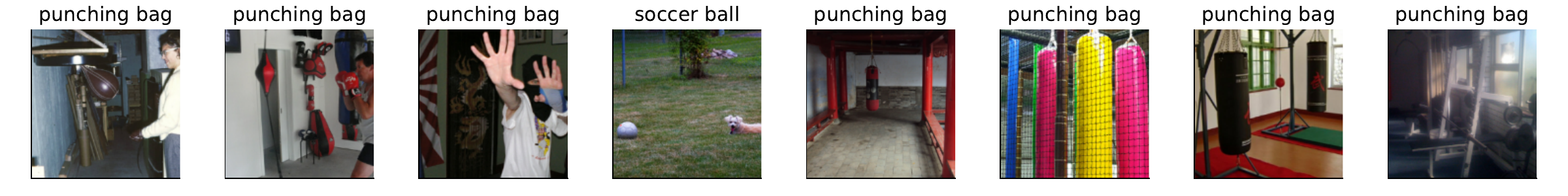}
\includegraphics[width=\linewidth]{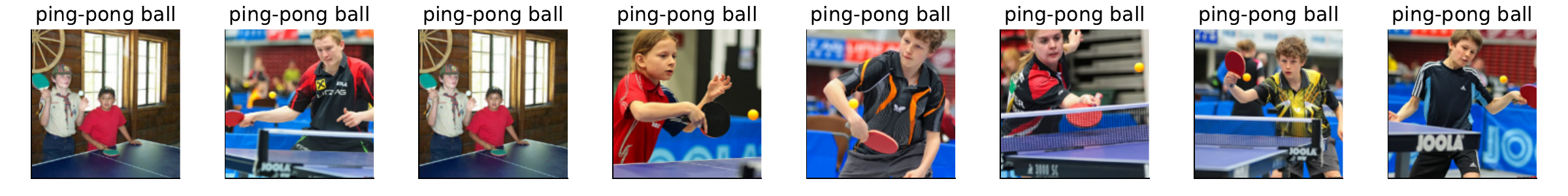}
\vskip -1em
\caption{Sample images from 4 clusters under `ball' uncovered in stage4\_block3 of ResNet-50 random seed 3.}
\end{figure}

\end{document}